%% file: main.tex
\documentclass{article}
\usepackage[T1]{fontenc}
\usepackage{iclr2027_conference,times}
\iclrfinalcopy
\usepackage{multirow}
\usepackage{booktabs}
\usepackage{arydshln}
\usepackage{graphicx}
\usepackage{amsmath}
\usepackage{xspace}
\usepackage{amssymb}
\usepackage{subcaption}
\usepackage{enumitem}
\usepackage{enumerate}
\usepackage{wrapfig}
\usepackage{caption}
\usepackage{tabularx}
\usepackage{needspace}
\usepackage{eso-pic}
\usepackage{microtype}
\usepackage{xcolor}
\usepackage{hyperref}
\usepackage{url}
\input{link_safety}
\makeatletter
\renewcommand{\@maketitle}{\vbox{\hsize\textwidth\centering
  {\LARGE\sc \@title\par}
  \vskip 0.2in
  {\@author\par}
  \vskip 0.25in minus 0.1in}}
\makeatother
\hypersetup{
    pdftitle={StereoPolicy: Improving Robotic Manipulation Policies via Stereo Perception},
    pdfauthor={Evans Han, Yunfan Jiang, Yingke Wang, Haoyue Xiao, Huang Huang, Jianwen Xie, Jiajun Wu, Li Fei-Fei, Ruohan Zhang},
    pdfsubject={},
    pdfkeywords={Robotic Perception, Stereo Vision, Robot Manipulation},
    colorlinks=true,
    linkcolor=blue!60!black,
    citecolor=blue!60!black,
    urlcolor=blue!60!black
}

\input{preamble}

\title{StereoPolicy: Improving Robotic Manipulation Policies via Stereo Perception}

\author{
{\bfseries Evans Han$^{1,2}$} \quad
{\bfseries Yunfan Jiang$^{1}$} \quad
{\bfseries Yingke Wang$^{1,*}$} \quad
{\bfseries Haoyue Xiao$^{1,*}$} \\
{\bfseries Huang Huang$^{1}$} \quad
{\bfseries Jianwen Xie$^{3}$} \quad
{\bfseries Jiajun Wu$^{1}$} \quad
{\bfseries Li Fei-Fei$^{1}$} \quad
{\bfseries Ruohan Zhang$^{1,2}$} \\
[0.5ex]
$^{1}$Stanford University \qquad
$^{2}$Northwestern University \qquad
$^{3}$Lambda, Inc
}

\begin{document}
\maketitle
\lhead{}
\renewcommand{\headrulewidth}{0pt}
\begingroup\makeatletter\let\thefootnote\relax\long\def\@makefntext#1{\noindent #1}\makeatother
\footnotetext{Correspondence to Evans Han \href{mailto:evanshan@stanford.edu}{\texttt{<evanshan@stanford.edu>}}.\\$^{*}$Denotes equal contribution.}\endgroup
\vspace{-6pt}

\input{sections/0_abstract}    
\input{sections/1_intro}

\input{sections/3_1_method}
\input{sections/4_experiment}

\input{sections/4_1_ablations}

\input{sections/2_related}
\input{sections/5_conclusion}

\bibliographystyle{iclr2027_conference}
\bibliography{main}

\clearpage
\input{sections/X_suppl}

\end{document}

%% file: link_safety.tex
\usepackage{etoolbox}
\usepackage{iftex}
\makeatletter
\ifXeTeX
  \protected\def\SP@linkoff{}
  \protected\def\SP@linkon{}
\else
  \ifPDFTeX
    \ifdefined\pdfrunninglinkoff
      \protected\def\SP@linkoff{\pdfrunninglinkoff}
      \protected\def\SP@linkon{\pdfrunninglinkon}
    \else
      \PackageError{stereopolicy-links}
        {pdfTeX is too old for running-link suppression}
        {Select a current TeX Live version in Overleaf.}
    \fi
  \else
    \ifLuaTeX
      \protected\def\SP@linkoff{\pdfextension linkstate 1 }
      \protected\def\SP@linkon{\pdfextension linkstate 0 }
    \else
      \PackageError{stereopolicy-links}
        {Unsupported compilation engine}
        {Use pdfLaTeX, XeLaTeX, or LuaLaTeX.}
    \fi
  \fi
\fi
\AtBeginDocument{%
  \patchcmd{\@outputpage}
    {\hb@xt@\textwidth{\@thehead}}
    {\hb@xt@\textwidth{\SP@linkoff\@thehead\SP@linkon}}
    {\typeout{StereoPolicy links: header protection installed.}}
    {\PackageError{stereopolicy-links}
       {Could not protect the page header}
       {The output routine has changed; inspect its header box.}}%
  \patchcmd{\@outputpage}
    {\hb@xt@\textwidth{\@thefoot}}
    {\hb@xt@\textwidth{\SP@linkoff\@thefoot\SP@linkon}}
    {\typeout{StereoPolicy links: footer protection installed.}}
    {\PackageError{stereopolicy-links}
       {Could not protect the page footer}
       {The output routine has changed; inspect its footer box.}}%
  \pretocmd{\iclrruler}
    {\SP@linkoff}
    {\typeout{StereoPolicy links: ruler-start protection installed.}}
    {\PackageError{stereopolicy-links}
       {Could not protect the start of the ICLR ruler}
       {Inspect the definition of \string\iclrruler.}}%
  \apptocmd{\iclrruler}
    {\SP@linkon}
    {\typeout{StereoPolicy links: ruler-end protection installed.}}
    {\PackageError{stereopolicy-links}
       {Could not protect the end of the ICLR ruler}
       {Inspect the definition of \string\iclrruler.}}%
}
\makeatother

%% file: preamble.tex
\newcommand{\algo}{StereoPolicy\xspace}
\newcommand{\algodp}{\textsc{StereoPolicy-DP}\xspace}
\newcommand{\algovla}{\textsc{StereoPolicy-VLA}\xspace}

%% file: sections/0_abstract.tex
\begin{abstract}
Recent advances in robot imitation learning have produced powerful visuomotor policies that manipulate diverse objects from visual inputs. However, monocular observations lack depth information, which is critical for precise manipulation in cluttered or geometrically complex scenes. Explicit depth maps and point clouds are often noisy and fragile in real-world manipulation. We introduce \textbf{StereoPolicy}, a visuomotor policy learning framework that directly leverages synchronized stereo image pairs to improve spatial perception without constructing explicit 3D representations. StereoPolicy processes each image with pretrained 2D vision encoders and fuses left-right features through a cross-attention-based Stereo Transformer, learning task-conditioned binocular representations. The framework integrates with diffusion-based and pretrained vision-language-action (VLA) policies, delivering consistent improvements over RGB, RGB-D, point cloud, and multi-view baselines across three simulation benchmarks
and seven real-robot tabletop and bimanual mobile manipulation tasks. Our results support task-trained binocular fusion as an effective way to adapt 2D visual representations for manipulation. 
The project page with demonstration videos and additional details is available at \url{https://stereopolicy.github.io}.
\end{abstract}

%% file: sections/1_intro.tex
\begin{figure}[h]
    \centering
    \vspace{-20pt}
    \includegraphics[width=0.98\linewidth]{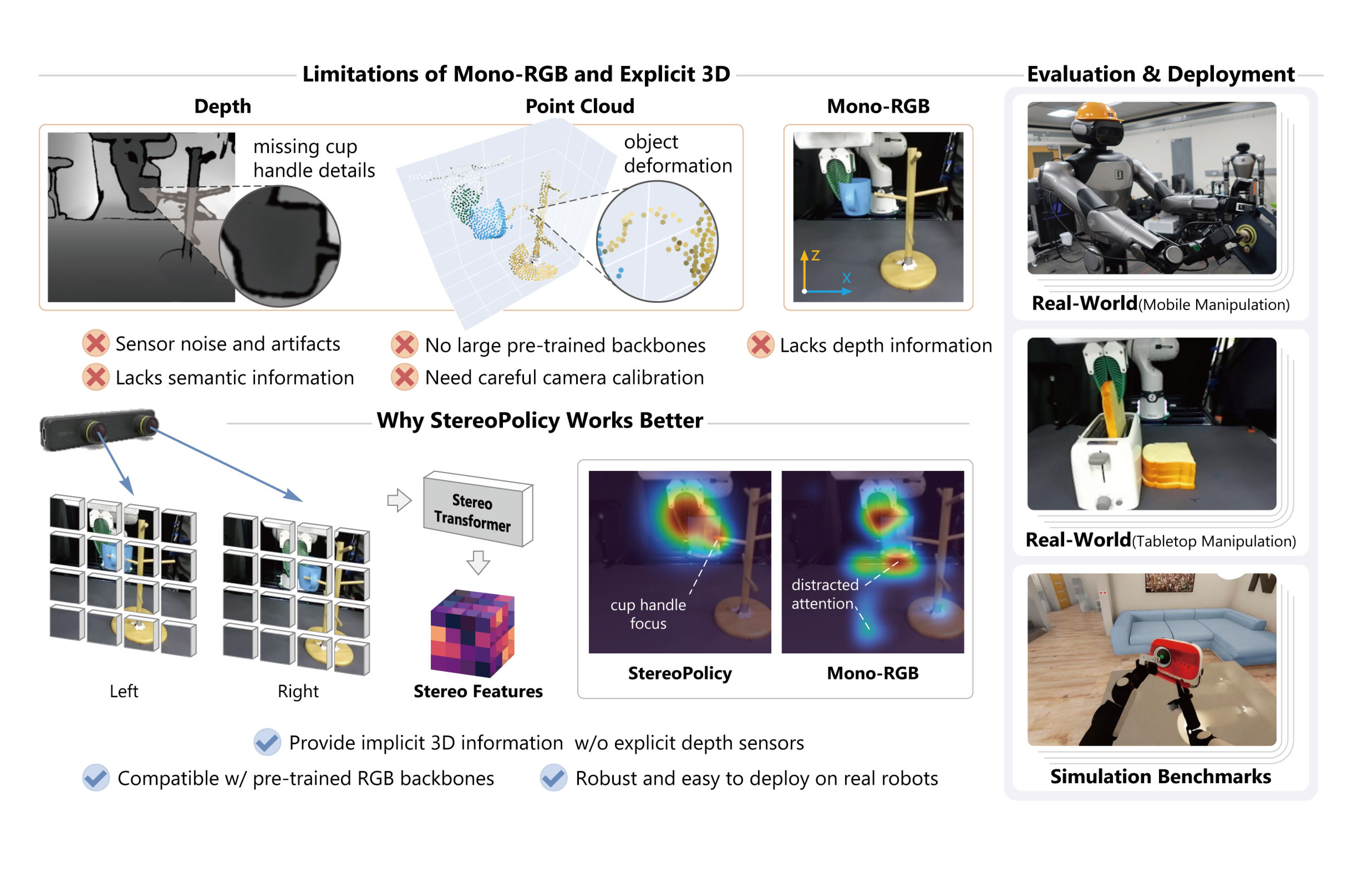}
    \vspace{-8pt}
    \caption{Compared with other visual modalities for robot learning, using stereo image pairs directly as policy input provides implicit 3D information without a depth sensor, works with pretrained 2D encoders, and is easy to deploy on real robots.}
    \label{fig:pull}
    \vspace{-12pt}
\end{figure}

\section{Introduction}
\label{sec:intro}

Large-scale robot learning has recently produced impressive visuomotor policies and vision-language-action (VLA) models that perform diverse manipulation tasks from visual inputs alone~\citep{levine2016endtoendtrainingdeepvisuomotor,zhu2018reinforcementimitationlearningdiverse,shridhar2021cliport,reed2022gato,jiang2022vima,brohan2022rt1,brohan2023rt2,dp,aloha,kim2024openvla,black2024_0,gr00tn1_2025}. These policies typically operate on monocular images processed by pretrained 2D vision encoders to extract semantic and spatial features for downstream action generation~\citep{resnet,radford2021learningtransferablevisualmodels,dosovitskiy2021imageworth16x16words}. While monocular inputs can be effective in controlled scenes, recovering accurate depth and geometry from a single view remains challenging. These capabilities are crucial for robots to operate in complex everyday environments~\citep{liu2024visual,uppal2024spin,yang2024equibotsim3equivariantdiffusionpolicy,jiang2025brs,sundaresan2025homerlearninginthewildmobile} and perform dexterous, fine-grained manipulation~\citep{qin2022dexpoint0,wang2024dexcap}.

Stereo vision instead naturally encodes 3D spatial structure through synchronized image pairs as shown in \autoref{fig:pull}. Extensive research in computer vision~\citep{doi:10.1126/science.968482,10.1007/978-3-030-58536-5_25,Poggi_2020_CVPR,Xu2020AANetAA,lipson2021raftstereomultilevelrecurrentfield,Li_2021_ICCV,Shen2021CFNetCA,Li_2022_CVPR,Weinzaepfel_2023_ICCV,Xu_2023_CVPR,wen2025foundationstereozeroshotstereomatching} has shown that stereo pairs significantly improve depth estimation and geometric understanding over monocular images, suggesting strong potential for robotic applications where 3D information is necessary.

However, prior work in robotics typically employs stereo images indirectly, by first constructing explicit 3D representations such as depth maps~\citep{shankar2021learnedstereodepthrobotic,Yang2023NeuralVM,uppal2024spin,liu2024visual} and point clouds~\citep{qin2022dexpoint0,Goyal2023RVTRV,ze20243d,jiang2024transic,yang2024equibotsim3equivariantdiffusionpolicy,jiang2025brs,sundaresan2025homerlearninginthewildmobile}, and then feeding these into control or planning modules. While effective in controlled settings, such approaches face practical challenges. First, explicit 3D representations are sensitive to sensor noise, calibration errors, and object properties, making high-quality reconstruction difficult to achieve reliably. Even with careful calibration, depth maps often miss fine semantic details, and point clouds suffer from object deformation, as visualized in \autoref{fig:pull}. Second, pretrained 3D vision encoders remain less mature than their 2D counterparts, largely due to the scarcity of large-scale 3D datasets~\citep{6248074,Menze_2015_CVPR,10.1007/978-3-319-11752-2_3} and scalable 3D backbone architectures~\citep{pointnet,qi2017pointnet,Thomas_2019_ICCV,Zhao2020PointT,jaegle2021perceiver,NEURIPS2022_d78ece66,wu2024pointtransformerv3simpler}. Third, robotic manipulation requires both high precision and low latency, whereas stereo pipelines that produce explicit disparity maps~\citep{wen2025foundationstereozeroshotstereomatching} incur substantial inference overhead from cost-volume construction and iterative refinement, posing challenges for real-time manipulation.

In this work, we take a different approach. Instead of relying on explicit 3D reconstruction, we design \algo, a framework that directly consumes synchronized stereo image pairs for action generation. Our method leverages 2D vision encoders pretrained on internet-scale data to extract features from each image independently, then fuses them into stereo-aware tokens through cross attention in a Stereo Transformer. The fusion module is trained through the action objective, producing task-conditioned binocular features without explicit depth supervision or reconstruction. 

Our main results can be summarized as follows:
\begin{itemize}[leftmargin=0pt, label={}, itemsep=0pt]
    \item \textbf{Performance.} We empirically demonstrate that \algo outperforms policies based on RGB, RGB-D, point cloud, and multi-view across diverse real and simulated manipulation tasks, spanning tabletop and bimanual mobile settings. It can also be integrated to improve pretrained VLA models.

    \item \textbf{Hardware Insights.} We systematically analyze key hardware factors including camera angles and stereo baseline-to-distance ratio, deriving practical guidelines for configuring stereo setups.

    \item \textbf{Model Design.} We comprehensively study stereo encoder design, comparing different backbone choices and identifying an effective architecture for robust stereo-based manipulation.
\end{itemize}

%% file: sections/3_1_method.tex
\section{Method}
We propose \algo, a stereo perception module to extract task-conditioned binocular features for robot policies (\S\ref{sec:stereopolicy}). The resulting representations can be seamlessly integrated into both diffusion policies and VLA models without modifying their backbone architectures. Below, we describe two models that form the conditions of our experiments (detailed in \S\ref{exps}).
\begin{itemize}[leftmargin=*]
    \item \algodp integrates a stereo encoder into diffusion policy, with the policy backbone trained from scratch on each benchmark task (\S\ref{sec:dp}). 
    \item \algovla combines the stereo encoder with a pretrained VLA model and then fine-tunes the whole system end-to-end on each downstream task (\S\ref{sec:vla}).
\end{itemize}

\begin{figure*}[t]
    \centering
    \raisebox{0.0128038343\linewidth}[0.4863895531\linewidth][0pt]{%
        \includegraphics[width=1\linewidth]{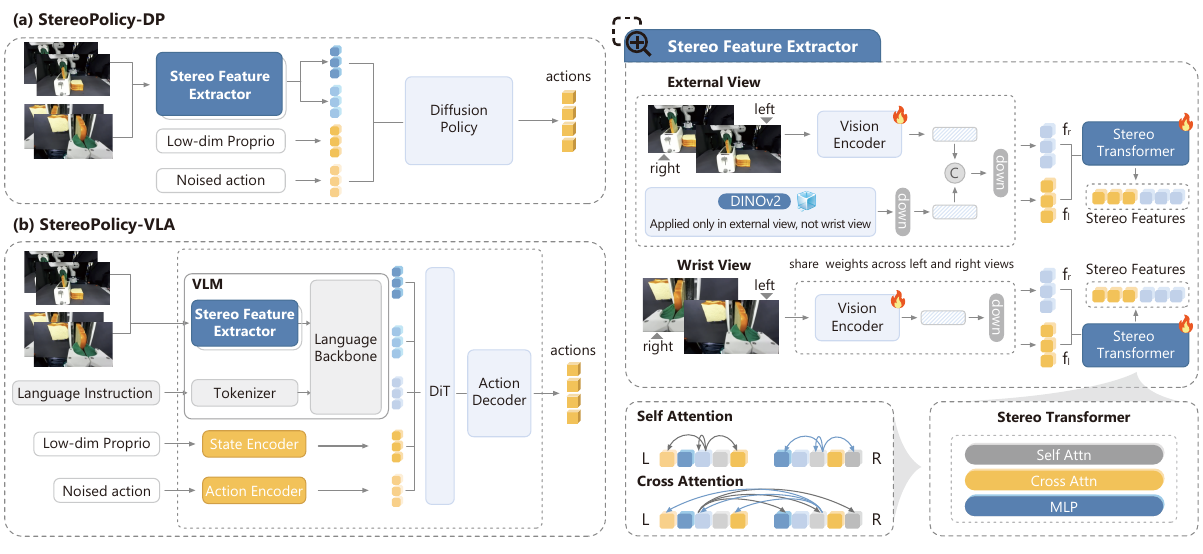}}
    
    \caption{\textbf{\algo Pipeline}. Stereo inputs are encoded by a vision backbone, fused with a Stereo Transformer, and applied to both diffusion policy training and fine-tuning VLA baselines.}
    \label{fig:pipeline}
    
\end{figure*}

\subsection{Stereo Feature Extraction}
\label{sec:stereopolicy}
Stereo vision has shown superior performance in 3D understanding, including depth estimation~\citep{wen2025foundationstereozeroshotstereomatching} and robot grasping~\citep{singh2024dextrah} in simulation environments. Building upon this observation, we design \algo that leverages stereo image pairs for robot manipulation. Formally, the policy consists of a stereo vision encoder and a policy backbone. The stereo vision encoder takes $N$ pairs of stereo images:
$\mathcal{I}_{t,i} = (I_{t,i}^{L}, I_{t,i}^{R}),$ where $(I_{t,i}^{L}, I_{t,i}^{R})$ denote the left and right images captured by a stereo camera setup from the $i$-th view at timestep $t$.
\paragraph{Stereo image encoding}
Each image is processed by a 2D vision encoder $V_i(\cdot)$ to produce single-view feature maps: 
${f_{t,i}^{L}, f_{t,i}^{R}}, \quad f^{\cdot} \in \mathbb{R}^{H \times W \times C}.$
Within each stereo pair, the left and right encoders share weights to preserve geometric consistency.
To enrich visual representations, we augment the 
image features with DINOv2~\citep{dinov2} embeddings, which provide strong monocular priors to disambiguate regions where epipolar correspondence is unreliable~\citep{wen2025foundationstereozeroshotstereomatching}:
\begin{equation}
\tilde{f}_{t,i}^{\cdot} = \phi\big([f_{t,i}^{\cdot} \Vert f_{t,i}^{\text{DINO}}]\big),
\end{equation} where $\phi(\cdot)$ is a linear projection aligning feature dimensions.
Empirically, we observe that DINOv2 improves external views but degrades wrist-view performance, likely due to domain mismatch with egocentric imagery. Therefore, DINOv2 is applied only to external views. We ablate the architecture without DINOv2, and the effect of applying it to wrist views, in \S\ref{sec:ablations}.
\paragraph{Stereo feature fusion}
To lift 2D image features with implicit depth cues, we adopt an alternating attention mechanism in Stereo Transformer: self-attention computes attention between pixels,  while cross attention computes attention between left and right images. We apply 2D RoPE~\citep{2drope} to cross-attention queries and keys using each token's image coordinates $(x,y)$ to encode spatial positions~\citep{singh2024dextrah}.
\begin{equation}
\forall i \in \{1,\dots,N\}: \quad 
z_{t,i} = \text{StereoTrans}_i\big(\tilde{f}_{t,i}^{L}, \tilde{f}_{t,i}^{R}\big) \in \mathbb{R}^{d_z},
\end{equation}

\paragraph{Multi-view aggregation}
Stereo fusion is performed independently per camera view. The resulting embeddings are then concatenated (denoted as $\Vert$) with low-dimensional proprioceptive features $s_t$ as the input to the policy backbone to predict robot actions:
\begin{equation}
z_{\text{obs}} = [\, z_{t,1} \Vert z_{t,2} \Vert \cdots \Vert z_{t,N} \Vert s_t \,].
\end{equation}

\begin{figure*}[t]
    \centering
    
    \includegraphics[width=1\linewidth]{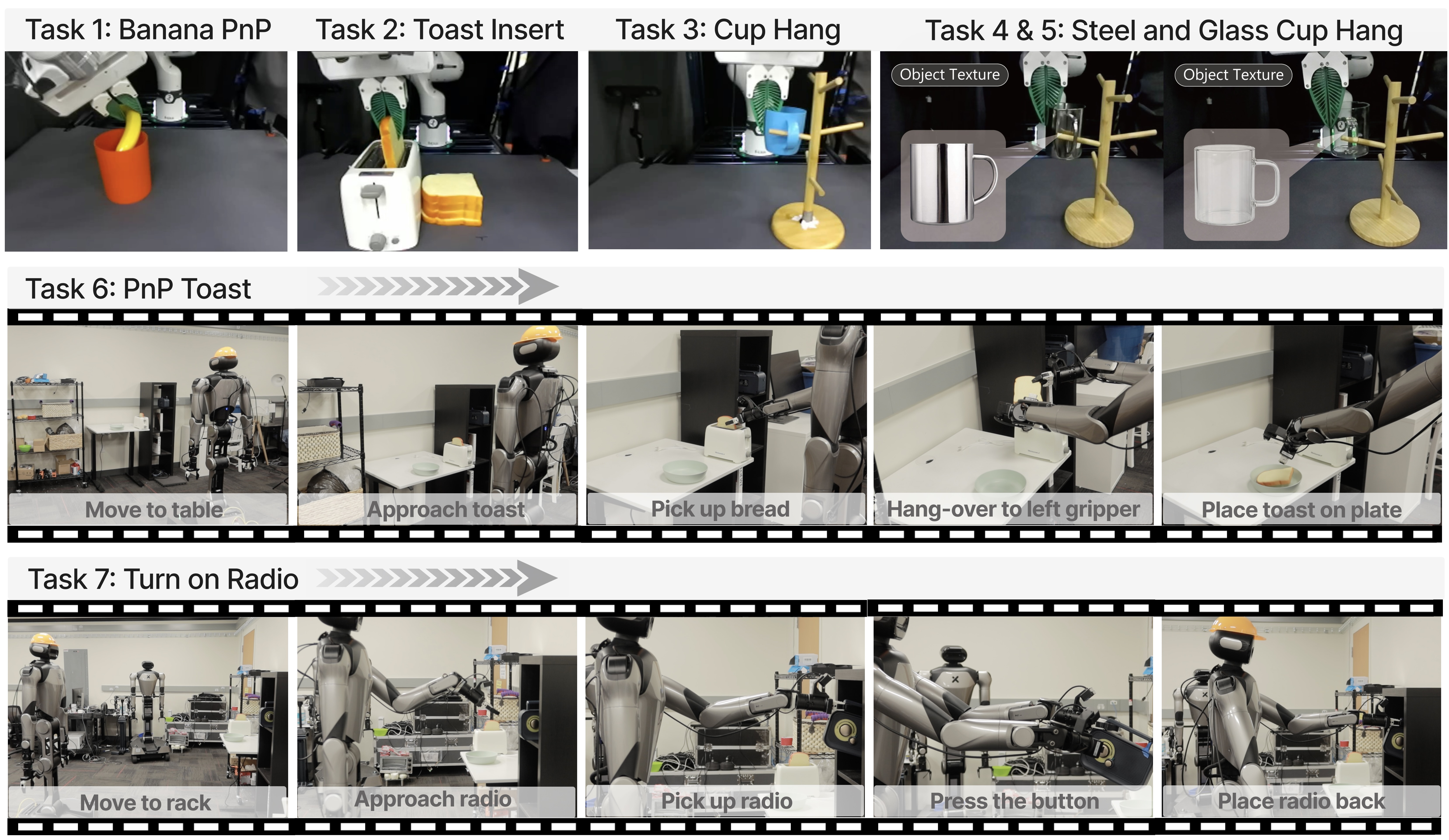}
    \caption{\textbf{Real-World Task Visualizations}. \textit{Top}: Five tabletop tasks, with the final state shown for each. Tasks 3--5 vary by cup material. \textit{Bottom}: Two mobile manipulation tasks.}
    
    \label{fig:real}
\end{figure*}

\subsection{\algodp: Diffusion Policy with \algo}
\label{sec:dp}
For imitation learning, we primarily adopt a diffusion policy~\citep{dp} as the policy backbone. 

The policy predicts a sequence of continuous actions $\mathbf{a}_{t:t+H-1} = [\mathbf{a}_t, \mathbf{a}_{t+1}, \ldots, \mathbf{a}_{t+H-1}]$ over a planning horizon $H$ based on $z_{\text{obs}}$. In \algodp, we inject stereo perception by conditioning the denoising network on the fused stereo embedding $z_{\text{obs}}$, which combines the fused stereo-view features. Following Diffusion Policy~\citep{dp}, we train the noise predictor conditioned on $z_{\text{obs}}$ with the denoising objective
\begin{equation}
\mathcal{L}_{\mathrm{DP}}
= \mathbb{E}\!\left[
\left\|\epsilon - \epsilon_{\theta}(x^{(\tau)}, \tau, z_{\text{obs}})\right\|_2^2
\right],
\end{equation}
where $\epsilon\sim\mathcal{N}(0,I)$ and $x^{(\tau)}$ is the demonstration action sequence corrupted by noise at diffusion step $\tau$. The expectation is over demonstration samples, diffusion steps, and Gaussian noise. The trainable vision encoder, Stereo Transformer, and policy backbone are optimized jointly, while \mbox{DINOv2} remains frozen. At inference, the policy generates actions by iteratively denoising an initial Gaussian sample. This design allows \algodp to enhance action generation with implicit stereo cues while preserving the original diffusion policy architecture and training objective.

\subsection{\algovla: Adapting Monocular VLA to Stereo Inputs}
\label{sec:vla}

Pretrained VLA models exhibit strong semantic understanding through VLM pretraining, but their depth reasoning is limited by monocular inputs. To improve spatial perception, we extend the visual input from monocular to stereo by introducing a lightweight stereo feature extractor on top of the pretrained vision encoder, without modifying the VLA backbone architecture.

Concretely, we replace the monocular visual embedding with a stereo-aware representation. For each stereo image pair, we extract features with the pretrained vision encoder from VLA, then inject the fused stereo representation from the Stereo Transformer into the VLA backbone to predict actions:
\begin{equation}
\mathbf{a}=\pi_\theta(z_{\text{obs}},x_\text{lang},s),
\end{equation}
where $\pi_\theta$ denotes the VLA policy (e.g., DiT), $z_{\text{obs}}$ is the fused stereo feature from StereoTrans, $x_{\text{lang}}$ is the language instruction, and $s$ denotes additional state.

%% file: sections/4_experiment.tex
\section{Experiments}\label{exps}
Our experiments were guided by the following research questions:
\begin{itemize}[leftmargin=0pt, label={}, itemsep=0pt]
\item $\mathbf{\mathcal{Q}1}$. How does \algo perform compared to monocular RGB, RGB-D, point cloud, and multi-view-based policies under the same policy backbone and training data?
\item $\mathbf{\mathcal{Q}2}$. Can \algo be readily combined with large pretrained policy models, such as vision-language-action (VLA) models, although these models are trained on monocular RGB data?
\item $\mathbf{\mathcal{Q}3}$. How do factors such as stereo baseline (distance between two cameras), distance to the objects, and camera angles affect \algo's performance? 
\item $\mathbf{\mathcal{Q}4}$. Can large-scale pretrained stereo models replace task-trained stereo fusion from \algo?
\end{itemize}

\begin{figure*}[t]
    \centering
    \includegraphics[width=1\linewidth]{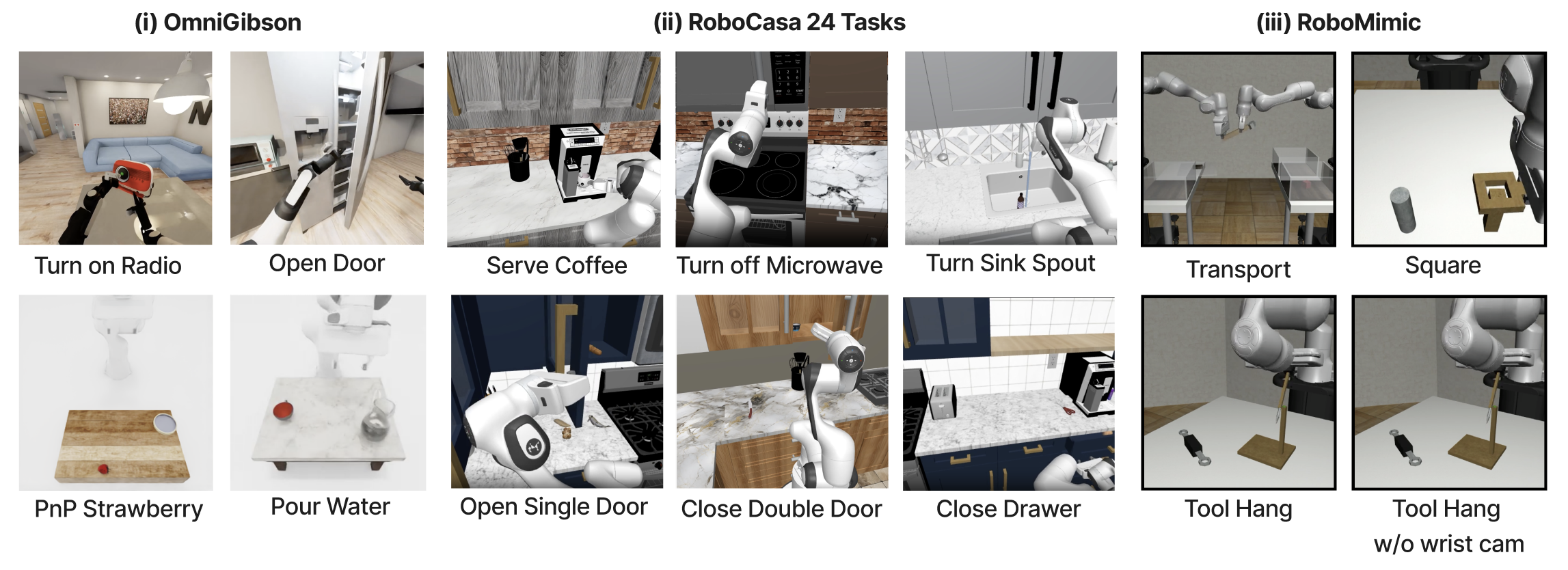}
    \caption{\textbf{Simulation Task Visualization}, from three benchmarks: \textsc{OmniGibson} (4 tasks), \textsc{RoboCasa} (24 tasks), and \textsc{Robomimic} (3 tasks).}
    \label{fig:task}
\end{figure*}

\subsection{Tasks and Environments}
We evaluate \algo on two robotic embodiments: a 7-DoF tabletop Franka arm and a 24-DoF Galaxea R1 Pro bimanual mobile manipulator with two 7-DoF arms, in both real-world and simulated settings. Task and data collection details are summarized in \autoref{app:tasks}.

\textbf{Real-World}
We design seven tasks spanning both tabletop and mobile manipulation (\autoref{fig:real}).

\textbf{Simulation} We include tasks from three widely used robotics benchmarks, including \textsc{Robomimic} \citep{robomimic}, \textsc{RoboCasa} \citep{robocasa}, and four customized tasks in \textsc{OmniGibson}~\citep{behavior}. \autoref{fig:task} visualizes tasks.  

\subsection{Implementation Details}

\paragraph{Baselines}
We compare \algodp against six baseline architectures: monocular RGB, multi-view RGB, RGB-D, point-clouds with PointNet~\citep{pointnet}. To compare \algodp with policies that more effectively utilize depth and point clouds, we also implement RGBD-3DDA~\citep{3dda} and PCD-DP3~\citep{ze20243d}. All baselines use the same policy backbones, image resolution, training budgets, and optimization hyperparameters where applicable.

To effectively evaluate \algovla, we adopt $\pi_{0.5}$~\citep{pi05} and GR00T-N1.5~\citep{gr00tn1_2025} as our baseline VLA frameworks and follow the same settings from the original implementation. The StereoPolicy architecture, baseline implementations, training and evaluation details are described in Appendix~\ref{app:implementation}.

\section{Results and Analysis}
\label{sec:results}

\subsection{Experimental Results}

\input{tables/real_tabletop}
\input{tables/benchmarks}

\input{tables/robocasa_avg}  

\paragraph{($\mathbf{\mathcal{Q}1}$) \algodp consistently outperforms monocular RGB, RGB-D, point cloud, and multi-view-based policies.} 
\autoref{tab:real_result} and \autoref{tab:benchmark} compare \algodp with standard visual modalities in real-robot and simulation respectively. Overall, \algodp presents strong performance in both settings. In particular, \algodp outperforms RGB across all tasks, highlighting the benefit of stereo inputs for manipulation. 
For less depth-sensitive tasks (e.g., \textit{Transport}), monocular RGB inputs already provide sufficient visual information for task success. For tasks involving occlusion and fine-grained 3D alignment (e.g., \textit{ToolHang} and \textit{PourWater}), \algodp shows clear improvements~\footnote{We present the inference latency of RGB, RGBD-3DDA and \algodp in Appendix~\ref{app:results}.}.

\begin{wrapfigure}{r}{0.48\linewidth}
    \centering
    \vspace{-4pt}
    \includegraphics[width=\linewidth]{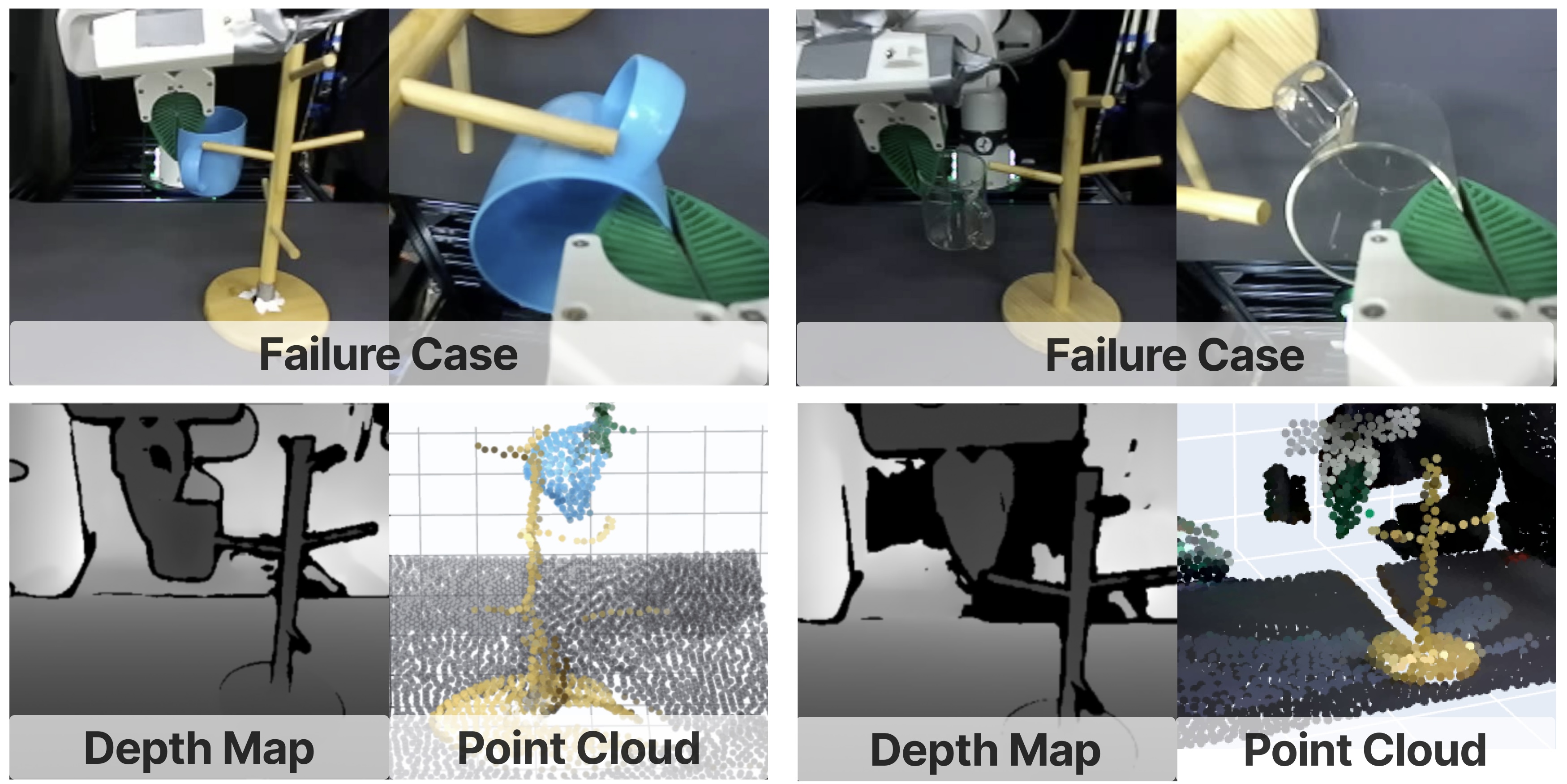}
    \caption{RGB-D and PCD are fragile in real environments. The glass cup is entirely missing.}
    \vspace{-10pt}
    \label{fig:failure}
\end{wrapfigure}
However, point cloud--based methods perform poorly overall: depth measurements are often noisy, leading to reconstructed point clouds with missing or distorted task-relevant geometry (see \autoref{fig:failure}). Even with access to clean depth in simulation, DP3 still underperforms on most benchmarks. We also observe point-cloud methods are fragile and heavily rely on careful pre-processing. Without careful cropping and sampling, their performance can easily drop to 0. We additionally discuss the failure reason of point-clouds policies in Appendix~\ref{app:pcd_failure}. While geometry-aware models such as RGBD-3DDA achieve competitive results in simulation due to perfect depth, their performance degrades notably on reflective and transparent objects 
(e.g., steel and glass cup hang), where depth sensing becomes unreliable; their reliance on calibrated depth sensors also limits robustness. In contrast, \algodp achieves higher success from image pairs in these evaluated settings, although transparent-object performance remains low in absolute terms. \algodp also outperforms RGBD-3DDA in most evaluated simulation settings, demonstrating the effectiveness of stereo inputs for manipulation without explicit depth or point-cloud reconstruction.

\textbf{\algodp outperforms multi-view-based policies.}
Our parameter-matched multi-view RGB baselines also use stereo images but concatenate their features without a Stereo Transformer, supporting the effectiveness of our stereo fusion. 
Therefore, the performance gain of \algodp is not solely from the additional view, but from the stereo fusion design.

\textbf{\algodp is more data-efficient.} In \textsc{OmniGibson} tasks, \algodp reaches higher SR with fewer demonstrations. Although \citet{peri2024pointcloudmodelsimprove} claims point-cloud inputs can yield higher sample efficiency compared to RGB/RGB-D, here we observe the opposite. 

\paragraph{($\mathbf{\mathcal{Q}2}$) \algo enhances pretrained VLA models, although these models are trained on monocular data.} We next examine \algovla, where \algo is incorporated into two state-of-the-art pretrained VLA models, $\pi_{0.5}$ and GR00T-N1.5, for fine-tuning. As summarized in \autoref{tab:robocasa_avg}, incorporating stereo input and stereo fusion module yields consistent improvements across both high-data (300 and 100 demos) and low-data (30 demos) settings. 
These gains, while modest, support our earlier claim: stereo features can provide geometric cues complementary to pretrained VLM representations, and fine-tuning allows these models to combine stereo cues with pretrained semantic features. We also compare \algovla with a VLA that directly takes the two stereo images as its multi-view input in Appendix Table~\ref{tab:review_vla}.

\autoref{fig:mobile} further evaluates \algovla on mobile manipulation. \algovla outperforms monocular baselines in both simulation and real-world settings. In particular, RGB policies fail on precise actions such as inserting the gripper into a radio handle and pressing buttons accurately (failure case visualization in Appendix), while \algovla succeeds.
 
\paragraph{($\mathbf{\mathcal{Q}3}$) \algodp is most effective when the baseline is approximately 10\% of the target object distance.} We vary the stereo baseline distance (2\,cm, 4\,cm, 6\,cm, 8\,cm, 10\,cm) and camera--object distance (0.6\,m--1.0\,m) while keeping other conditions fixed. Results show that performance is governed not by either factor alone, but by their ratio: 
\begin{equation}
    r = \frac{\text{stereo baseline}}{\text{camera-object distance}}
\end{equation}
which determines the effective disparity. \algodp achieves the best performance when $r$ lies in the range of $r \in [0.09, 0.13]$, as shown in \autoref{fig:baseline}. 

\begin{figure}[t]
    \centering
    \begin{subfigure}[t]{0.34\linewidth}
        \centering
        \raisebox{0.4cm}{%
            \includegraphics[width=\linewidth]{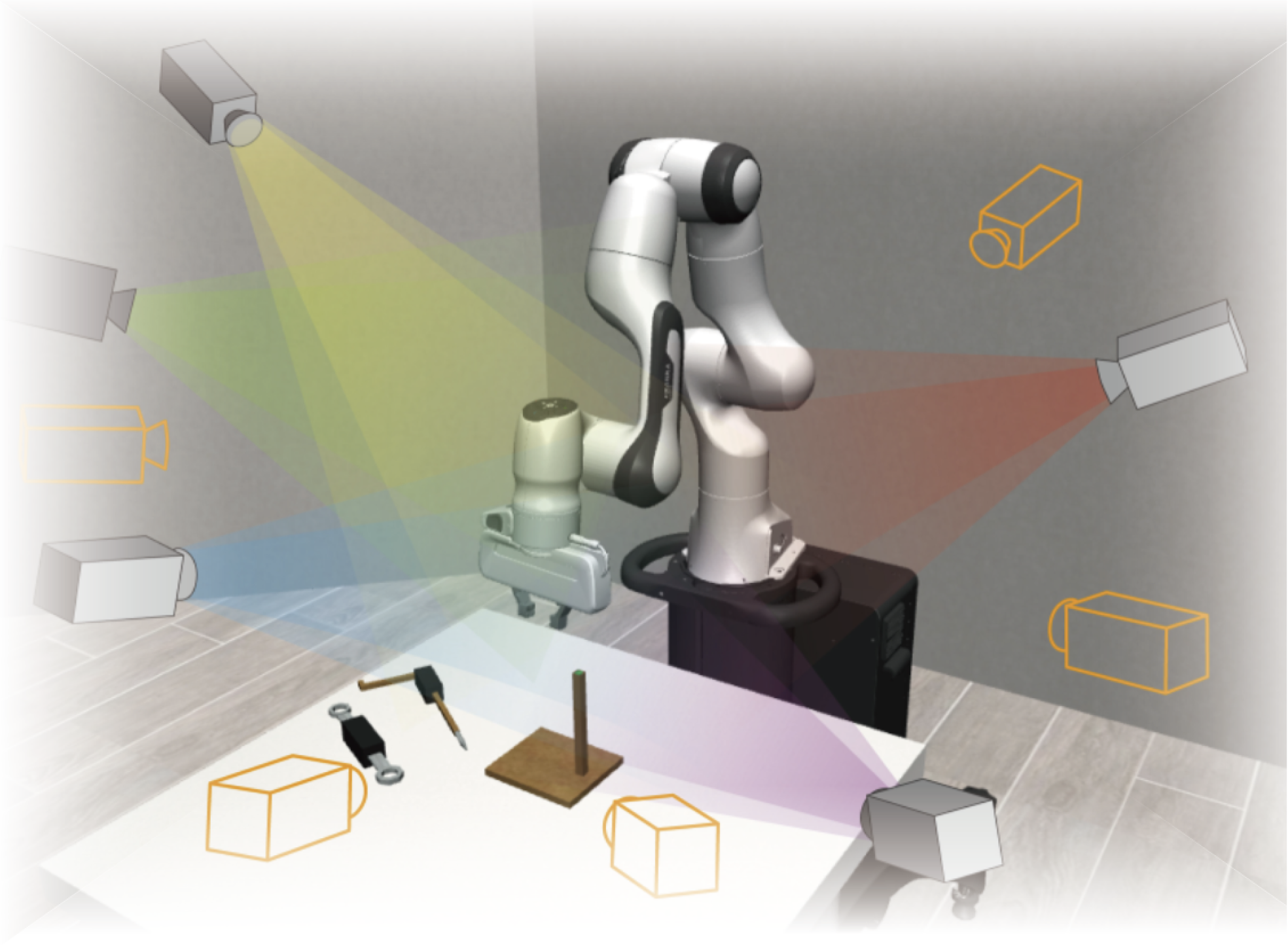}%
        }
        \caption{\textbf{Camera Angle Visualization}. From left to right: \textit{sideview left, agentview left, frontview, agentview right, and sideview right}.}
        \label{fig:camera_angle}
    \end{subfigure}
    \hfill
    \begin{subfigure}[t]{0.64\linewidth}
        \centering
        \includegraphics[width=\linewidth]{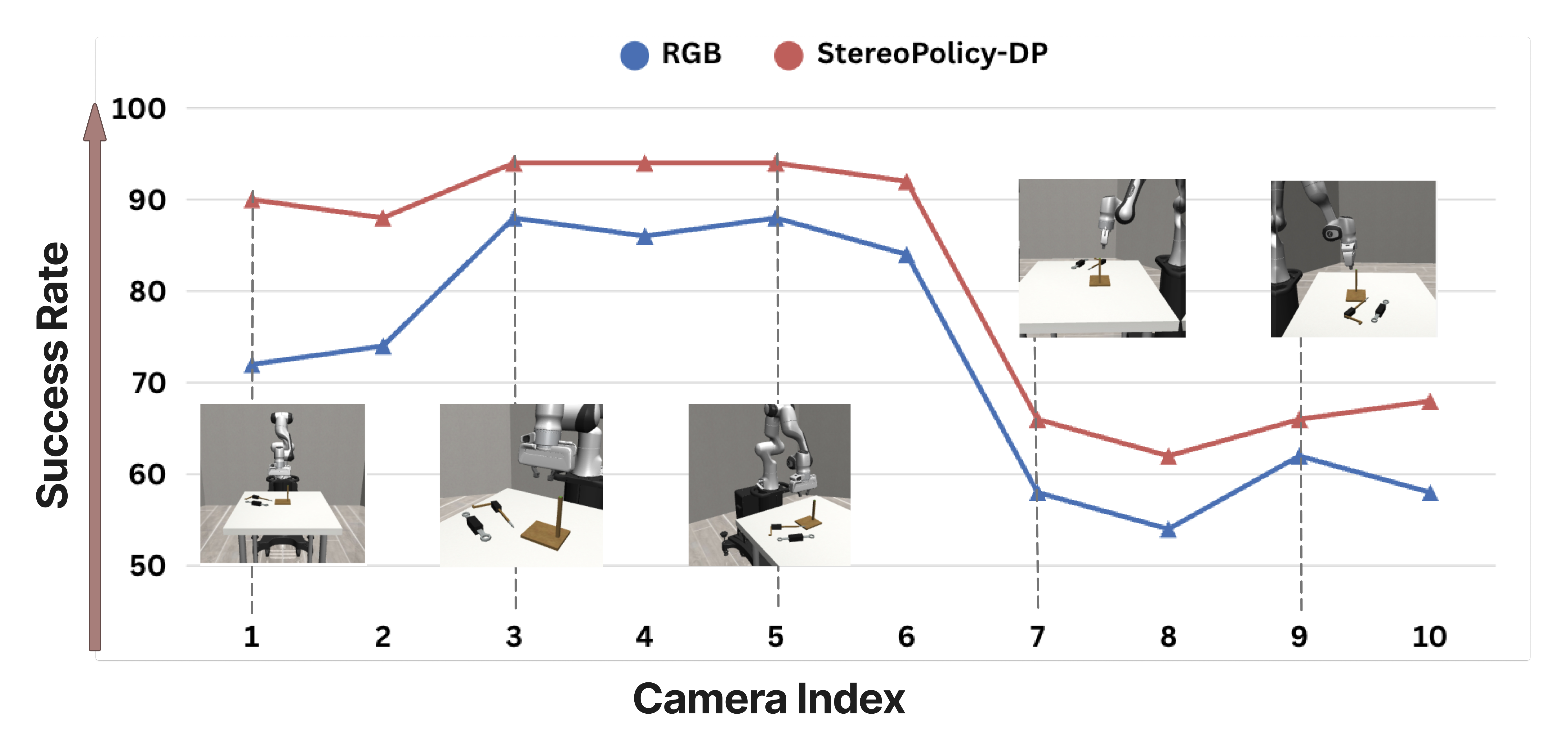}
        \caption{\textbf{SR across 10 Different Camera Angle Views} over \textsc{ToolHang} task for RGB policy and \algodp. \textit{\algodp is more robust towards changing camera angles than RGB baseline}. Appendix~\ref{app:hardware} provides full mapping of camera indices.}
        \label{fig:angle}
    \end{subfigure}
    \caption{Performance of \algodp across different camera angles.}
    \vspace{-10pt}
\end{figure}

Within this range, disparity is strong enough for reliable depth estimation while preserving sufficient overlap between left and right views. 
For typical tabletop setups (0.6--0.8m camera-object distance), a 6 cm baseline tends to perform better than smaller or larger ones.

When $r$ becomes too small (e.g., 2\,cm baseline or $>0.9$\,m distance), performance drops due to weak disparity and poor depth discrimination. Conversely, excessively large $r$ values (e.g., 10\,cm baseline at 0.6\,m) reduce view overlap and cause geometric inconsistency across views.
In conclusion, stereo policies are most effective when the baseline is approximately 10\% of the target distance. 
This ratio maintains an optimal, balanced trade-off between disparity and overlap. It allows \algodp to extract stable geometric features and improves manipulation accuracy across viewpoints.

\textbf{In addition, \algodp consistently outperforms RGB-based policies across different camera angles.} We evaluate 10 camera angles on the \textsc{ToolHang} task, visualized in \autoref{fig:camera_angle}; \autoref{fig:angle} shows the corresponding views and performance. The largest gains appear in front-facing views (\textit{frontview}, +18\%; \textit{frontview-high}, +14\%), where stereo disparity provides complementary depth cues. Oblique views such as \textit{agentview-left/right} and their elevated counterparts also benefit, as stereo helps resolve partial occlusions and spatial correspondence. Side views show smaller gains due to limited overlap and stronger self-occlusion. Overall, stereo is most helpful when monocular cues are geometrically ambiguous, and it helps in all ten tested views.

\input{tables/targeted_comparison}

\paragraph{($\mathbf{\mathcal{Q}4}$) Pre-trained Stereo features cannot replace task-adapted stereo fusion.} Table~\ref{tab:targeted_comparison} compares \algodp{} with two ways of using a pretrained stereo model on ToolHang with 100 demonstrations, keeping the diffusion-policy backbone fixed. First, following StereoVLA~\citep{deng2025stereovla}, we replace our trainable encoder and Stereo Transformer with frozen FoundationStereo features. Success
decreases from 94\% to 76\%, indicating that a generic pretrained stereo representation does not
directly substitute for task-adapted fusion in this setting.  Second, we replace the depth source in RGBD and 3DDA policies by depth inferred with (Fast) FoundationStereo. However, they incur substantially higher end-to-end latency while their success rate degrades greatly, likely because depth errors introduced by FS affect the reconstructed 3D coordinates. On the real robot, the stale observation caused by this latency becomes a more significant practical problem (Appendix~\ref{app:real_fs}).

%% file: tables/real_tabletop.tex
\begin{table}
    \vspace{0pt}
    \centering

    \resizebox{\linewidth}{!}{%
    \begin{tabular}{lcccccc}
    \toprule
    \textbf{Method} & 
    \shortstack{\textsc{Banana}\\\textsc{PnP}} & \shortstack{\textsc{Toast}\\\textsc{Insert}} & \shortstack{\textsc{Plastic}\\\textsc{Cup Hang}} & \shortstack{\textsc{Steel}\\\textsc{Cup Hang}} & \shortstack{\textsc{Glass}\\\textsc{Cup Hang}} & AVG SR (\%) \\
    \midrule
    RGB & 12/20 &  8/20 &  12/20 &  10/20 &  1/20 & 43.0\% \\
    RGB-D             & 14/20 & 8/20 & 11/20 & 8/20  & 0/20 & 41.0\%\\
    RGBD-3DDA      & 13/20 & 9/20 & 13/20 & 10/20 & 0/20 & 45.0\%\\
    PCD-PointNet    & 7/20  & 0/20 & 5/20  & 2/20  & 0/20 &  14.0\%\\
    PCD-DP3        & 11/20 & 3/20 & 8/20  & 5/20  & 0/20 & 27.0\%\\
    Multi-View & 13/20 & 9/20 & 13/20 & 9/20 & 1/20 & 45.0\%\\
    \algodp           & \textbf{16/20} & \textbf{12/20} & \textbf{15/20} & \textbf{13/20} & \textbf{3/20} & \textbf{59.0\%}\\
    \bottomrule
    \end{tabular}
    }
    
\caption{\textbf{Real-World Tabletop Task Performance}. \algodp consistently outperforms other visual modalities. PCD performs worst in all real tasks; both RGB-D and PCD fail on glass cup hang tasks, demonstrating that explicit 3D methods are sensitive to object texture in practice. }
\label{tab:real_result}
\end{table}

%% file: tables/benchmarks.tex
\begingroup
\setlength{\tabcolsep}{4.2pt}
\renewcommand{\arraystretch}{1}
\newcommand{\simcell}[2]{\ensuremath{#1_{\scriptscriptstyle #2}}}
\newcommand{\simcellb}[2]{\ensuremath{\mathbf{#1}_{\scriptscriptstyle #2}}}
\begin{table}[t]
\centering

\resizebox{1\linewidth}{!}{%
\begin{tabular}{ccccccccccccccc}
\toprule
\textbf{Benchmark} &
\multicolumn{8}{c}{\textsc{OmniGibson}}& \multicolumn{6}{c}{\textsc{Robomimic}}  \\
\cmidrule(lr){2-9} \cmidrule(lr){10-15}
\textbf{Task} &
\multicolumn{2}{c}{\textsc{Strawberry}} &
\multicolumn{2}{c}{\textsc{PourWater}} &
\multicolumn{2}{c}{\textsc{OpenDoor}} &
\multicolumn{2}{c}{\textsc{TurnOnRadio}} &
\multicolumn{2}{c}{\textsc{ToolHang}} &
\multicolumn{2}{c}{\textsc{Square}} &
\multicolumn{2}{c}{\textsc{Transport}}  \\
\cmidrule(lr){2-3} \cmidrule(lr){4-5} \cmidrule(lr){6-7} \cmidrule(lr){8-9}
\cmidrule(lr){10-11} \cmidrule(lr){12-13} \cmidrule(lr){14-15}
\textbf{\# Demos} &
100 & 200 & 100 & 200 & 200 & 300 & 200 & 300 & 100 & 200 & 100 & 200 & 100 & 200  \\
\midrule

RGB
& \simcell{62}{3.3} & \simcell{86}{1.6}
& \simcell{15}{4.1} & \simcell{50}{2.8}
& \simcell{30}{3.3} & \simcell{80}{1.6}
& \simcell{46}{2.8} & \simcell{74}{2.8}
& \simcell{53}{3.4} & \simcell{90}{1.6}
& \simcell{76}{1.9} & \simcell{98}{0.9}
& \simcell{92}{1.9} & \simcell{94}{1.6} \\

RGB-D &
\simcell{63}{3.8} &
\simcell{85}{4.7} &
\simcell{16}{1.6} &
\simcell{52}{0.0} &
\simcell{31}{2.5} &
\simcell{80}{4.3} &
\simcell{47}{0.9} &
\simcell{73}{3.4} &
\simcell{56}{0.0} &
\simcell{88}{2.5} &
\simcell{79}{1.9} &
\simcell{92}{0.0} &
\simcellb{94}{1.6} &
\simcell{94}{1.6} \\

RGBD-3DDA &
\simcell{74}{4.3} &
\simcell{93}{1.9} &
\simcell{26}{0.0} &
\simcell{61}{3.4} &
\simcell{48}{3.3} &
\simcellb{100}{0.0} &
\simcell{46}{4.3} & 
\simcell{75}{4.1} &
\simcell{84}{1.6} &
\simcell{92}{2.8} &
\simcell{83}{1.9} &
\simcell{97}{2.5} &
\simcellb{94}{1.6} &
\simcellb{96}{1.6} \\

PCD-DP3 &
\simcell{45}{3.4} &
\simcell{63}{2.5} &
\simcell{3}{4.7} &
\simcell{31}{0.9} &
\simcell{30}{1.6} &
\simcell{69}{4.1} &
\simcell{35}{1.9} & 
\simcell{64}{4.9} &
\simcell{40}{1.6} &
\simcell{76}{0.9} &
\simcell{69}{2.5} &
\simcell{88}{1.6} &
\simcell{63}{2.5} &
\simcell{72}{2.8} \\

Multi-View
& \simcell{70}{3.3} & \simcell{91}{2.5}
& \simcell{24}{3.4} & \simcell{56}{3.3}
& \simcell{35}{4.3} & \simcell{82}{2.8}
& \simcell{47}{3.4} & \simcell{77}{2.5}
& \simcell{64}{2.8} & \simcell{92}{1.9}
& \simcell{80}{1.6} & \simcell{96}{0.9}
& \simcell{92}{1.9} & \simcell{94}{1.6} \\

\algodp &
\simcellb{82}{2.8} &
\simcellb{100}{0.0} &
\simcellb{34}{1.6} &
\simcellb{70}{2.5} &
\simcellb{57}{1.9} &
\textbf{\simcellb{100}{0.0}} &
\simcellb{55}{3.4} & 
\simcellb{82}{4.3} &
\simcellb{94}{1.9} &
\simcellb{96}{2.8} &
\simcellb{88}{0.9} &
\simcellb{100}{0.0} &
\simcellb{94}{0.0} &
\simcellb{96}{1.6} \\
\bottomrule
\end{tabular}%
}

\caption{\textbf{Simulation Task Performance of Diffusion Policies across Visual Modalities}. Stereo input consistently improves performance, especially under low-data regime. We report average scores and standard deviations across three seed runs. All numbers are reported in percentage (\%).}
\label{tab:benchmark}
\end{table}
\endgroup

%% file: tables/robocasa_avg.tex
\begin{figure}[t]
\centering

\begin{minipage}{0.5\linewidth}
\centering
\includegraphics[width=\linewidth]{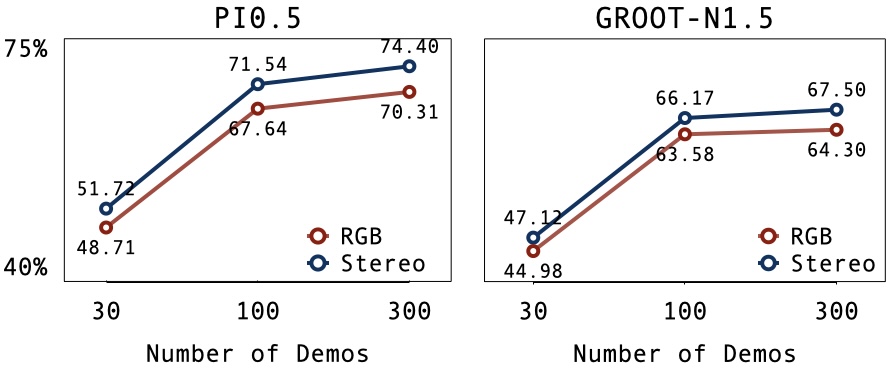}
\caption{Average \algovla Performance on \textsc{RoboCasa-Kitchen} 24 Tasks.}
\label{tab:robocasa_avg}
\end{minipage}
\hfill
\begin{minipage}{0.48\linewidth}
\centering
\includegraphics[width=\linewidth]{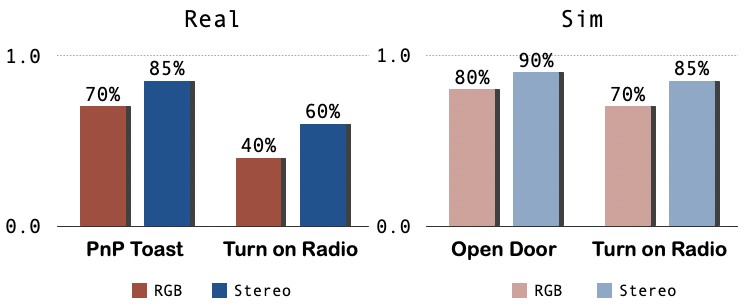}
\caption{\algovla ($\pi_{0.5}$) Performance on bimanual mobile manipulation tasks in both real-world and simulation.}
\label{fig:mobile}

\end{minipage}
\vspace{-10pt}
\end{figure}

%% file: tables/targeted_comparison.tex
\begin{figure}[h]
\centering
\begin{minipage}[t]{0.43\linewidth}
\vspace{0pt}
\centering
\includegraphics[width=\linewidth]{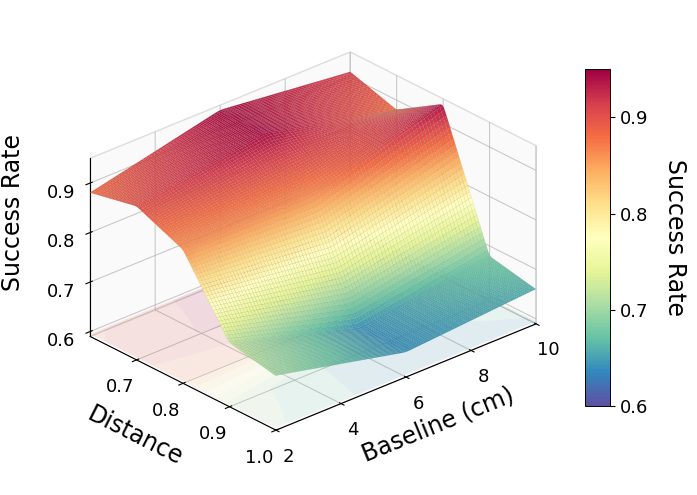}
\captionof{figure}{Effect of stereo baseline and camera--object distance on task success rate.}
\label{fig:baseline}
\end{minipage}%
\hfill
\begin{minipage}[t]{0.54\linewidth}
\vspace{-4pt}
\centering
\resizebox{\linewidth}{!}{%
\begin{tabular}{lcc}
\toprule
\textbf{Method} & \textbf{SR (\%)} & \textbf{Latency (ms)} \\
\midrule
\algodp & $94_{1.9}$ & 698 \\
\midrule
\multicolumn{3}{l}{\textit{A. Pretrained stereo features}} \\
Frozen FS features & $76_{4.3}$ & 740 \\
\midrule
\multicolumn{3}{l}{\textit{B. Stereo-derived geometry}} \\
FS depth + RGB-D & $44_{4.3}$ & 995 \\
FastFS depth + RGB-D & $38_{3.3}$ & 792 \\
FS depth + 3DDA & $28_{2.8}$ & 1008 \\
FastFS depth + 3DDA & $22_{2.8}$ & 777 \\
\bottomrule
\end{tabular}%
}
\vspace{-4pt}
\captionof{table}{Comparison with pre-trained stereo foundation models on ToolHang (100 demo). FS denotes FoundationStereo. Subscripts are std over three seeds.}
\label{tab:targeted_comparison}
\end{minipage}
\vspace{-10pt}
\end{figure}

%% file: sections/4_1_ablations.tex
\subsection{Ablations}
\label{sec:ablations}

We examine which part of \algodp contributes most to the success rate. Figure~\ref{fig:ablation} removes one component at a time (numbers in Appendix Table~\ref{tab:no_dino_sim}). The largest drop comes from removing binocular information. 
When the left image is fed to both branches, success falls from 84.0\% to 62.7\%.
DINOv2 helps, but it is not the source of the stereo gain. 
\begin{wrapfigure}{r}{0.48\linewidth}
    \centering
    \vspace{-10pt}
    \includegraphics[width=\linewidth]{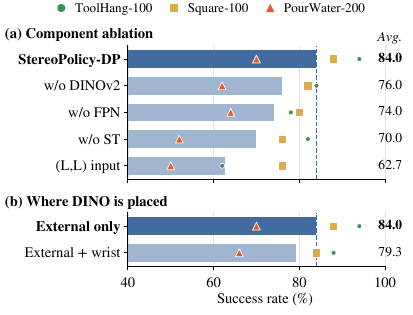}
    \vspace{-16pt}
    \caption{\textbf{Component Ablation}. ST denotes StereoTransformer.}
    \vspace{-15pt}
    \label{fig:ablation}
\end{wrapfigure}
Without it, the policy still reaches 76\% on average and stays ahead of both single-image controls on every task. Removing the Stereo Transformer hurts significantly on most tasks. On these tasks, the second image adds little unless the two views are fused through cross-attention. The FPN features are also important for stereo feature extraction. We also explore how to appropriately use DINO features. Applying DINO to the wrist images hurts all tasks' success, likely due to a domain mismatch. Therefore, we apply DINO only on the external view. Together with the additional ablations in Appendix~\ref{app:ablations}, these results indicate that the gains come mainly from stereo fusion, suggesting \algodp learns stereo geometry.

%% file: sections/2_related.tex
\section{Related Work}
\label{app:related_work}

\textbf{Stereo Vision}
Stereo matching recovers 3D structure from rectified image pairs. Modern methods estimate disparity or depth with cost volumes, multi-scale features, and recurrent refinement~\citep{wen2025foundationstereozeroshotstereomatching, min2025stextsuperscript2mtextsuperscript2scalablestereomatching, bartolomei2025stereoanywhererobustzeroshot, xu2025igeviterativemultirangegeometry, lipson2021raftstereomultilevelrecurrentfield, chang2018pyramidstereomatchingnetwork}, and recent models transfer zero-shot to challenging conditions~\citep{wen2025foundationstereozeroshotstereomatching, wang2025robustereorobustzeroshotstereo}. In robotics, stereo-based manipulation systems often use such models to reconstruct depth first~\citep{shankar2021learnedstereodepthrobotic,bai2025cleardepthenhancedstereoperception,kollar2021simnetenablingrobustunknown}, while stereo grasping policies can learn directly from images~\citep{singh2024dextrah,lum2024dextrah}. Concurrent with our work, StereoVLA~\citep{deng2025stereovla} combines pretrained FoundationStereo features with semantic features and adds auxiliary depth and camera-parameter supervision. StereoPolicy differs from both lines of work: it learns task-conditioned binocular features directly through the action objective, on top of generic pretrained 2D encoders, without explicit disparity estimation, stereo-specific pretraining, or depth supervision.

\textbf{3D-aware Policies and VLAs}
Many policies take explicit 3D input. DP3~\citep{ze20243d} and FP3~\citep{yang2025fp3} encode reconstructed point clouds, Act3D~\citep{gervet2023act3d} and 3D Diffuser Actor~\citep{3dda} lift 2D features into 3D with depth and camera calibration, and RVT~\citep{Goyal2023RVTRV} and RVT-2~\citep{goyal2024rvt2} re-render point clouds into virtual views. Similar ideas have been applied to VLAs via point-cloud features~\citep{li2025pointvla}, projected point-cloud images~\citep{li2025bridgevla}, and 3D position encodings from monocular depth estimates~\citep{qu2025spatialvla}. These methods depend on depth sensing and calibration or on estimated depth, whereas StereoPolicy learns geometric cues end-to-end from a rectified stereo pair.
 
\textbf{Multi-view Policy Learning}
Recent multi-view policies go beyond concatenating per-view features and model cross-view structure explicitly, e.g., by rendering a fused RGB-D point cloud into virtual views for a video diffusion model~\citep{li2026mvvdp}, or by reconstructing a shared 3D Gaussian field from several agents' cameras~\citep{wang2025gaudp}. StereoPolicy addresses a different and complementary problem: it exploits dense correspondence between two nearly identical rectified images from a single viewpoint without explicit depth or pcd reconstruction. Because stereo fusion occurs within each camera rig, it could be combined with cross-view aggregation as in those works. 

%% file: sections/5_conclusion.tex
\section{Conclusion}
We presented StereoPolicy, a visuomotor learning framework that directly consumes synchronized stereo image pairs instead of explicit 3D reconstructions. By combining pretrained 2D vision encoders with a cross-attention-based stereo fusion module, StereoPolicy improves over monocular RGB, RGB-D, point-cloud, and multi-view baselines across both simulated and real-world settings, spanning tabletop and mobile manipulation tasks. Our analysis further identifies practical design choices for stereo-based policies, including camera baseline, viewpoint, vision backbone, and fusion architecture. StereoPolicy also improves pretrained VLA models, suggesting that stereo cues can complement large-scale 2D visual representations for robot learning without explicit 3D supervision.

%% file: sections/X_suppl.tex
\appendix

\noindent{\Large\bfseries Appendix}
\vspace{0.75em}

\section{Task and Data Collection Details}
\label{app:tasks}

\paragraph{Simulation Tasks}
\begin{itemize}[itemsep=2pt, parsep=0pt, topsep=2pt, partopsep=0pt, leftmargin=*]
    \item \textsc{Robomimic}:  
    We exclude \textit{Lift} and \textit{Can} since the success rate has already saturated in the original study, and include three more challenging tasks \textit{ToolHang}, \textit{Square}, and \textit{Transport}. 
    \item \textsc{RoboCasa-Kitchen}:
    We use 24 kitchen tasks from this benchmark, and the main purpose is to evaluate the pretrained VLA models ($\pi_{0.5}$ and GR00T-N1.5).
    \item \textsc{OmniGibson} is a high-fidelity physics simulation environment that models a large variety of object interactions in realistic home-scale scenes, enabling long-horizon mobile manipulation tasks grounded in everyday human activities. We design four tasks and collect demonstrations using cuRobo~\citep{curobo_report23}.
\end{itemize}

\begin{figure}[!htb]
    \centering
    \includegraphics[width=1\linewidth]{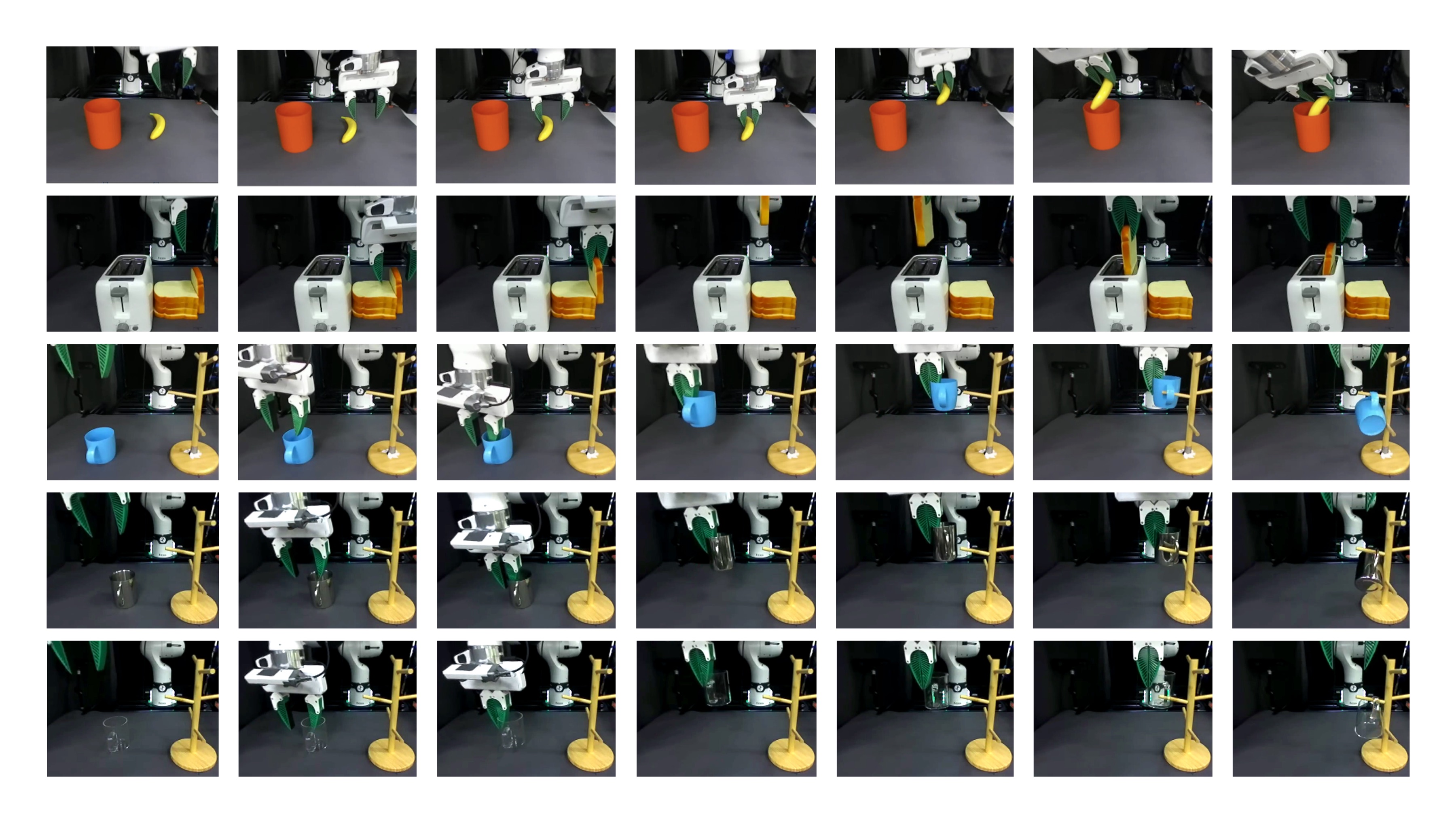}
    \caption{Example trajectories of real-world tabletop tasks.}
    \label{fig:traj}
\end{figure}

\paragraph{Real Robot Tasks.} We design five real-world tabletop manipulation tasks and collect 200 demonstrations per task. Example demonstration trajectories are shown in \autoref{fig:traj}.

\begin{itemize}[itemsep=2pt, parsep=0pt, topsep=2pt, partopsep=0pt, leftmargin=*]
    \item \textsc{Banana PnP}: the robot picks up a banana and inserts it into a container.
    \item \textsc{Toast Insert}: the robot picks up a slice of toast and inserts it into the toaster.
    \item \textsc{Plastic Cup Hang}: the robot picks up a plastic mug and hangs it on the holder. 
    \item \textsc{Steel Cup Hang}: the robot picks up a steel mug and hangs it on the holder. 
    \item \textsc{Glass Cup Hang}: the robot picks up a glass mug and hangs it on the holder.
\end{itemize}
For the following bimanual mobile manipulation tasks, we collect 75 demonstrations per task. 
\begin{itemize}[itemsep=2pt, parsep=0pt, topsep=2pt, partopsep=0pt, leftmargin=*]
    \item \textsc{Turn on Radio}: The R1Pro robot approaches the shelf and picks up the radio, then uses its left gripper to press the radio's button, and puts the radio back on the shelf. 
    \item \textsc{Toast PnP}: The R1Pro robot approaches the table, picks up the toast from the plate, hands it over to the left gripper and places it on the shelf. 
\end{itemize}

\subsection{Real-Robot Hardware Setup}
To further validate the real-world applicability, we deploy the stereo policy on a 7-DoF Franka Emika robot arm. For visual perception, we utilize the dual camera setup: a ZED Mini provides an external view, and a wrist-mounted ZED Mini captures a close-range view. For the bimanual R1 Pro setup, each wrist has its own stereo camera. 

\section{Stereo Hardware Configurations}
\label{app:hardware}

\subsection{Stereo Baselines}
In simulation, the default stereo baselines are $6$\,cm for external cameras and $2$\,cm for wrist cameras. The real-world ZED Mini cameras have a $6.3$\,cm baseline.

\Needspace{6\baselineskip}
\subsection{Test-Time Baseline Transfer}
\label{app:camera_transfer}
The baseline--distance sweep in the main paper compares camera configurations. Here, we hold the trained policy fixed: training uses a 6\,cm baseline, and evaluation uses 4, 6, or 8\,cm at a fixed camera--object distance. Success decreases from 94\% at the training baseline to 86\% at 4\,cm and 88\% at 8\,cm (Table~\ref{tab:baseline_transfer_existing}). The 8- and 6-point decreases show tolerance to moderate baseline changes without retraining, with the best performance at the training configuration.
\begin{table}[htbp]
\centering\small
\begin{tabular}{lrrr}\toprule
Test baseline & 4 cm & 6 cm & 8 cm\\\midrule
Success rate (\%) & $86_{1.6}$ & $94_{1.9}$ & $88_{1.6}$\\\bottomrule
\end{tabular}
\caption{ToolHang SR after training at a 6\,cm baseline and changing only the baseline at test time. Subscripts are standard deviations over three seeds.}
\label{tab:baseline_transfer_existing}
\end{table}

\subsection{Stereo Views across Different Baselines and Distances}
\begin{figure}[!htb]
    \centering
    \includegraphics[width=1\linewidth]{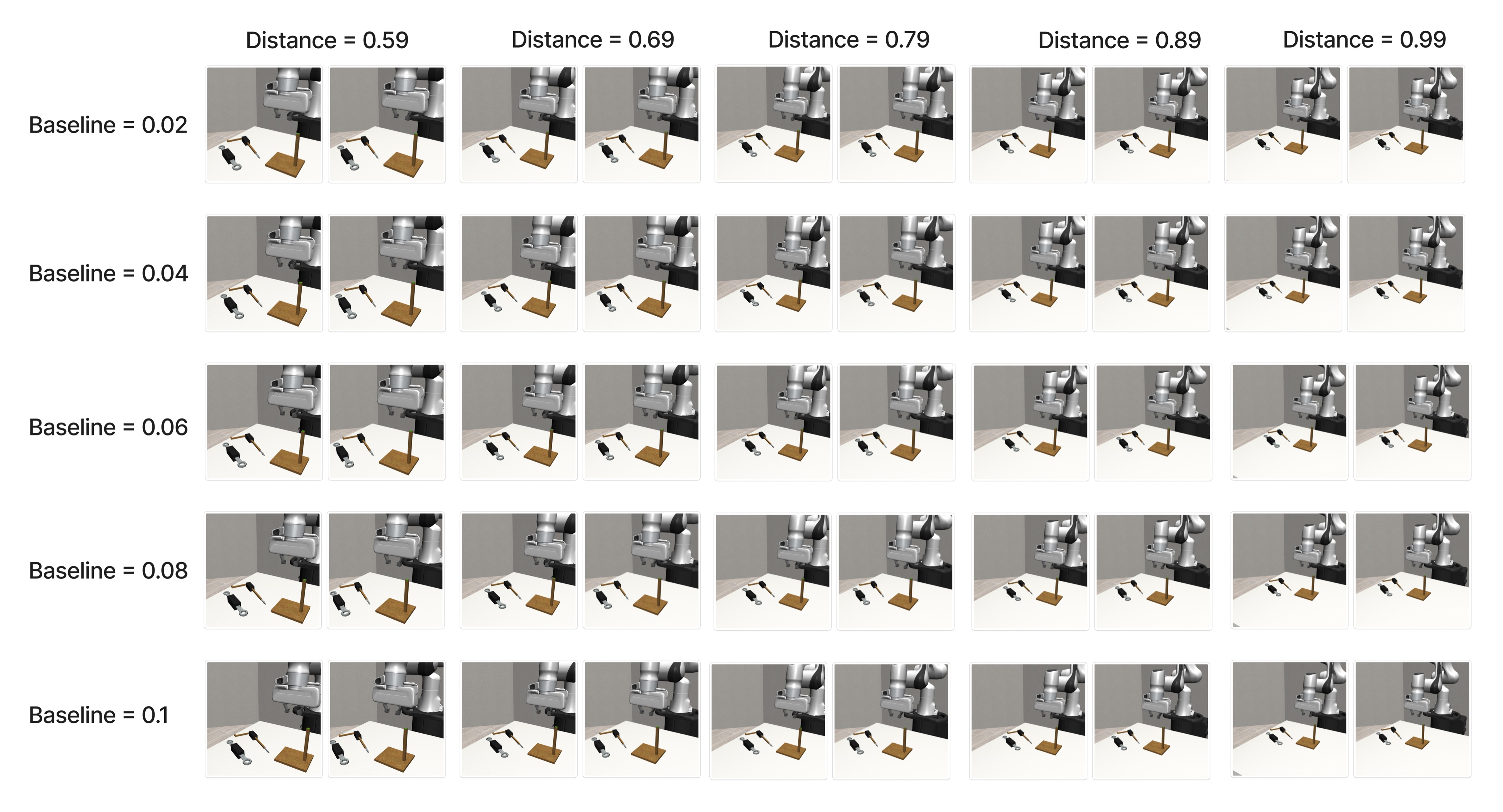}
    \caption{Stereo camera views across different baselines and distances. Baseline indicates the distance between stereo cameras; distance indicates the camera-object distance.}
    \label{fig:all_baseline}
\end{figure}

In \S\ref{sec:results} ($\mathbf{\mathcal{Q}3}$), 
we examine how stereo baseline and camera-object distance affect \algo. In \autoref{fig:all_baseline}, we visualize all stereo camera views across baselines ranging from 0.02\,m to 0.10\,m and object distances ranging from 0.59\,m to 0.99\,m that are used in the experiments. 

\subsection{Camera Angle Mappings}
\begin{figure}[!htb]
    \centering
    \includegraphics[width=1\linewidth]{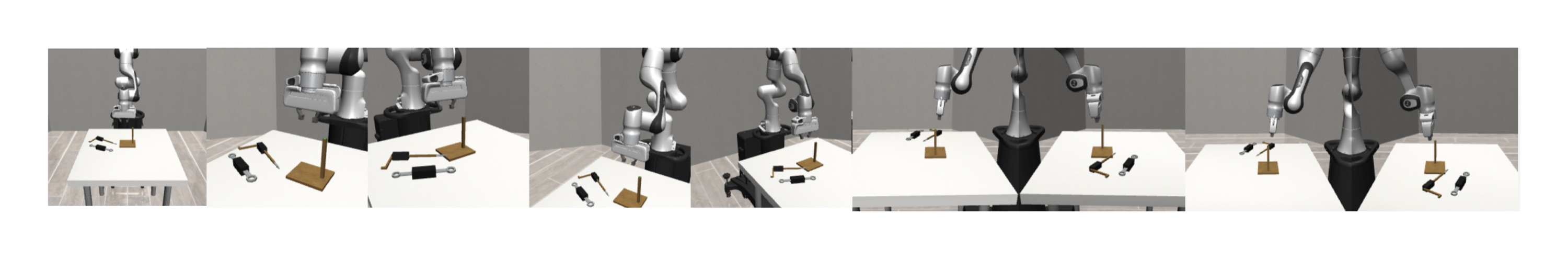}
    \caption{Visualization of the camera views. Experimental results are shown in \autoref{fig:angle}. }
    
    \label{fig:all_angle}
\end{figure}

As reported in
\S\ref{sec:results}  ($\mathbf{\mathcal{Q}3}$), we conduct experiments on the \textsc{ToolHang} task across 10 camera views. 
\autoref{fig:all_angle} shows the 10 views, which are \textit{frontview, frontview-high, agentview-left, agentview-right, agentview-righthigh, agentview-lefthigh, sideview-left, sideview-right, sideview-lefthigh, sideview-righthigh} from left to right. 

\section{Implementation Details}
\label{app:implementation}

\subsection{\algodp Training}

Each image is encoded by a ResNet18 with an FPN, with weights shared between the left and right images. For external cameras, frozen DINOv2 features are downsampled by a $4\times4$ convolution with stride 4, concatenated with the CNN features, and projected to the token dimension by an MLP. The Stereo Transformer has 2 layers and 8 attention heads. As described in \S\ref{sec:stereopolicy}, each layer applies self-attention within each image, cross-attention from two images, and an MLP block, with 2D RoPE applied to queries and keys. The output tokens are projected by an MLP to a 128-dimensional view feature. Wrist cameras use the same encoder without DINOv2.

For all diffusion policies, we use a UNet~\citep{ronneberger2015unetconvolutionalnetworksbiomedical} as the policy backbone and replace the BatchNorm layers in the UNet with GroupNorm~\citep{wu2018groupnormalization} for stable training~\citep{dp}.  We train the diffusion policy backbones from scratch. For RGB policies, we use images from the left cameras as input. All models are trained with a batch size of 64 and a learning rate of 1e-4. The observation horizon is set to 2 timesteps, while the action prediction horizon is set to 16 steps.  For OmniGibson tasks, we train for 1,000 epochs, while Robomimic tasks are trained for 500 epochs following common practice in prior work. 

\subsection{\algovla Training}

We fine-tune the pretrained \textsc{GR00T-N1.5} checkpoint on RoboCasa tasks for 60K gradient steps, maintaining the original training protocol as closely as possible to ensure comparability. Learning rate is set to 1e-4. Across all settings, optimization hyperparameters follow the corresponding baseline configurations to isolate the effect of our stereo architecture. 

For $\pi0.5$, we use the pretrained checkpoints to reproduce fine-tuned performance on RoboCasa-Kitchen. We train it for 80K steps, both with a global batch size of 128. We set the learning rate to 2.5e-5 with cosine decay to 2.5e-6 and 1K warmup steps. At
inference, we use an action horizon $H = 16$ and execute all actions without replanning. Both training uses 8 H100 GPUs. 

\subsection{Baseline Implementations}
\label{app:baseline_impl}

\begin{itemize}[itemsep=2pt, parsep=0pt, topsep=2pt, partopsep=0pt, leftmargin=*]
\item{\textbf{Monocular RGB}.}
Following Robomimic~\citep{robomimic} and Diffusion Policy~\citep{dp}, at each timestep $t$, the sensor inputs include monocular RGB images from each view $i$, $I_{t,i} \in \mathbb{R}^{H \times W \times 3}$. For a fair comparison, we augment the RGB policy with the same vision encoder and frozen external-view DINOv2 features as \algodp plus a self-attention Transformer, whose capacity is adjusted to approximately match our trainable parameter count. This comparison tests whether increased model capacity accounts for the improvement. 

\item{\textbf{Multi-view RGB}}
Stereo images from each view are passed to the same vision encoders as \algodp respectively, and the resulting image features are concatenated before the diffusion policy. The architecture and parameters are matching \algodp. To match the architecture and parameters of \algodp, we retain the simple multi-view fusion strategy: features from the two images are concatenated without token-level left--right attention, followed by an enlarged projection/MLP whose capacity is adjusted to match StereoPolicy. 

\item{\textbf{RGB-D}.}
Each view provides an RGB image and an aligned depth map, enabling explicit spatial structure for downstream processing. Depth maps $\in \mathbb{R}^{H \times W \times 1}$ are repeated to 3 channels and processed by a separate ResNet18 encoder. The resulting RGB and depth features are concatenated.

\item{\textbf{RGBD-3DDA}.} We follow the same setup as in the original paper~\citep{3dda}, use ResNet18 and FPN to extract multi-scale RGB features and reconstruct point-clouds from depth. 

\item{\textbf{PCD-PointNet}.}
We reconstruct and fuse multi-view point clouds from RGB-D observations, followed by downsampling them to 4096 points using farthest point sampling (FPS)~\citep{pointnet}.

\item{\textbf{PCD-DP3}.} For point-cloud inputs, we reconstruct and fuse multi-view point clouds from RGB-D observations, downsample them to 4096 points using farthest point sampling (FPS)~\citep{pointnet}, and encode them using the DP3 encoder~\citep{ze20243d}.

\item \textbf{Frozen FoundationStereo features (StereoVLA-style).} Following StereoVLA~\citep{deng2025stereovla}, we replace our trainable vision encoder and StereoTransformer with StereoVLA’s encoder using frozen FoundationStereo (FS) features under the same backbone
\item \textbf{Stereo-estimated depth.} FoundationStereo or Fast-FoundationStereo~\citep{wen2026fastfoundationstereo} estimates depth from the same rectified stereo pair, and this depth replaces the simulator depth in the RGB-D and RGBD-3DDA baselines above. Everything else in those baselines is unchanged.
\end{itemize}

\subsection{Evaluation} 
For simulation tasks, we perform 100 rollouts at every 50 training epochs for Robomimic tasks, and every 250 training epochs for OmniGibson tasks. The highest success rate achieved across training is reported with three different seeds.
For real-robot evaluation, we report the average success rate over 20 trials, with randomized initial poses. 

\section{Additional Ablations}
\label{app:ablations}

\subsection{Vision Encoder}
To select vision encoders that mostly utilize stereo images, table~\ref{tab:encoder_sweep} ablates image encoder variants without DINOv2 on ToolHang with 100 demonstrations. ResNet18+FPN reaches the highest success rate (84\%). These results support the benefit of extracting multi-scale feature from stereo image pairs, and ResNet18+FPN acts as a competitive encoder in this setting. 

\input{tables/no_dino_sim}

\subsection{Attention Design}
We hold the rest of the model architecture, wrist observations, training data, and optimization budget fixed, and change only how the stereo images are fused. Both branches receive identical augmentation for duplicated $(L,L)$ input. 
Joint attention concatenates the left and right tokens into a single sequence and applies standard self-attention. The Stereo Transformer alternates within-image and cross-image attention. The attention variants share tokenization, feature width, and positional conventions. Results are mean success rates across three training seeds.
 
\begin{table}[htbp]
\centering\small
\begin{tabular}{llccc}\toprule
Fusion & Views & ToolHang-100 & Square-100 & PourWater-200\\\midrule
Joint attention & $L,L$ & $64_{1.6}$ & $78_{1.6}$ & $42_{2.8}$\\
Joint attention & $L,R$ & $82_{1.6}$ & $82_{1.6}$ & $60_{0.0}$\\
Stereo Transformer & $L,L$ & $62_{1.6}$ & $76_{1.6}$ & $50_{3.3}$\\
Stereo Transformer & $L,R$ & $94_{1.9}$ & $88_{0.9}$ & $70_{2.5}$\\
\bottomrule
\end{tabular}
\caption{Matched fusion success rates (\%). Task suffixes give demonstration counts. The Stereo Transformer rows correspond to \algodp and its $(L,L)$ variant in Table~\ref{tab:no_dino_sim}. Subscripts are standard deviations over three seeds.}
\label{tab:review_fusion}
\end{table}
 
When both branches see the same image, the fusion designs behave almost identically on ToolHang and Square. PourWater is the exception. Even with duplicated input, the Stereo Transformer reaches 50\%, compared with 42\% for joint attention
, so part of its advantage on this task comes from the architecture itself rather than from stereo.
The picture changes once the true right image is provided. The Stereo Transformer rises from 62\% to 94\% on ToolHang and from 76\% to 88\% on Square, whereas joint attention reaches only 82\% on both tasks
. On PourWater, joint attention benefits from the right image about as much as the Stereo Transformer does, and the Stereo Transformer keeps its lead at 70\%. Overall, the Stereo Transformer makes better use of the second view than the other fusion designs, and the difference is clearest on ToolHang and Square.

\subsection{Pre-trained Stereo Foundation Models on Real Robot}
\label{app:real_fs}

\paragraph{Real robot with FoundationStereo}
On real-world BananaPnP, FoundationStereo+RGBD succeeds in 1/20 trials, compared with 16/20 for StereoPolicy. The depth-based deployment exhibited significant stale observations and delayed actions. This comparison highlights the importance of the complete perception--control pipeline: its low success reflects the combined effects of depth estimation and execution timing, rather than isolating either factor.

\subsection{Geometric Information}
\label{app:review_geometry}
We freeze the visual representation before proprioceptive input and fit equal-capacity linear probes to camera-frame gripper--target relative depth. Feature dimensionality and probe training data are fixed. Evaluation holds out complete trajectories and spatial layouts.

\begin{table}[htbp]
\centering\small
\begin{tabular}{lrr}\toprule
& \multicolumn{1}{c}{ToolHang} & \multicolumn{1}{c}{PourWater}\\
Representation & MAE (cm)  & MAE (cm) \\\midrule
Joint attention, $L,L$ & 2.8  & 3.6 \\
Joint attention, $L,R$ & 1.9  & 2.5 \\
Stereo Transformer, $L,L$ & 2.7  & 3.5\\
Stereo Transformer, $L,R$, before training & 3.1  & 4.0 \\
Stereo Transformer, $L,R$, after training & 1.1  & 1.5 \\
\bottomrule
\end{tabular}
\caption{Relative-depth mean absolute error from frozen features.}
\label{tab:review_probe}
\end{table}

After action training, the stereo representation reduces depth error from 3.1 to 1.1\,cm on ToolHang and from 4.0 to 1.5\,cm on PourWater. At fixed stereo input, its errors are 42\% and 40\% lower than joint attention. The duplicated-input control does not reproduce these gains. These probes identify accessible geometric information.

\subsection{Discussion on Point Clouds Policy Performance}
\label{app:pcd_failure}
Our point-clouds preprocessing follows most standard preprocessing steps. Point-cloud baselines use the same camera views as the other modalities. Calibrated RGB-D observations are fused in a common frame, cropped to remove the background and retain the workspace, and downsampled by farthest-point sampling to 4096 points. Increasing the count to 10K did not improve performance in the evaluated setting and increased inference cost. This observation does not support point density as the main cause of the performance gap. 

Especially, one bottleneck of low-performance of PCD is instead real-world PCD quality:  \autoref{fig:pull} and \autoref{fig:failure} show missing or deformed task-relevant geometry, which is often happening in point clouds reconstruction. Second, Our real task ToastInsert is especially challenging as successful insertion requires precise 3D alignment with a narrow slot, which is depth-sensitive, the bad-quality point clouds will significantly affect the policy performance. 

\subsection{\algovla Performance}
We compare monocular input, native two-image input, and StereoPolicy fusion using the same pretrained checkpoints and fine-tuning budgets. Evaluation averages tasks equally across the 24 RoboCasa tasks, using three seeds and 100 test episodes per task and seed. 

\begin{table}[t]
\centering
\small
\begin{minipage}[t]{0.60\linewidth}
\vspace{0pt}
\centering
\resizebox{\linewidth}{!}{%
\begin{tabular}{llrrr}
\toprule
 & & \multicolumn{3}{c}{\# Demos} \\
\cmidrule(lr){3-5}
Backbone & Input & 30 & 100 & 300 \\
\midrule
$\pi_{0.5}$ & Monocular & $48.71_{0.7}$ & $67.64_{1.0}$ & $70.31_{0.6}$ \\
$\pi_{0.5}$ & Two-image & $49.01_{0.1}$ & $68.24_{0.7}$ & $70.81_{1.8}$ \\
$\pi_{0.5}$ & \algovla & $\mathbf{51.72}_{0.5}$ & $\mathbf{71.54}_{1.0}$ & $\mathbf{74.40}_{0.5}$ \\
\midrule
GR00T-N1.5 & Monocular & $44.98_{0.4}$ & $63.58_{0.6}$ & $64.30_{1.1}$ \\
GR00T-N1.5 & Two-image & $45.67_{1.6}$ & $65.24_{0.7}$ & $65.31_{0.7}$ \\
GR00T-N1.5 & \algovla & $\mathbf{47.12}_{0.9}$ & $\mathbf{66.17}_{1.2}$ & $\mathbf{67.50}_{1.0}$ \\
\bottomrule
\end{tabular}%
}
\caption{Matched VLA macro-average success (\%) on 24 RoboCasa tasks. Two-image: the backbone's native multi-view input with the left and right images. Subscripts are standard deviations over three seeds.}
\label{tab:review_vla}
\end{minipage}%
\hfill
\begin{minipage}[t]{0.37\linewidth}
\vspace{0pt}
\centering
\resizebox{\linewidth}{!}{%
\begin{tabular}{lc}
\toprule
 & Latency (ms) \\
Method & p50 / p95 \\
\midrule
RGB-DP & 632 / 674 \\
Joint attention & 695 / 731 \\
StereoPolicy & 698 / 736 \\
FS + 3DDA & 1008 / 1060 \\
FastFS + 3DDA & 777 / 819 \\
\bottomrule
\end{tabular}%
}
\caption{End-to-end inference latency (median / 95th percentile) on one H100 GPU.}
\label{tab:review_latency}
\end{minipage}
\vspace{-10pt}
\end{table}

Native two-image input improves both backbones over monocular input. StereoPolicy further provides improvements over native two-image. These gains identify a benefit from the fusion design beyond supplying a second image.
\subsection{Overhead, computational cost}
\label{app:results}
StereoPolicy uses 112M parameters, of which 10M belong to the Stereo Transformer, and the reported compute is 128 GFLOPs, peak memory 1.4G. Inference is profiled on one H100 with batch size one, observation horizon two, prediction horizon sixteen, action dimension seven, and 100 denoising steps. Timing excludes 100 warm-up calls and summarizes 1,000 synchronized calls. Stage medians and end-to-end median/95th-percentile latency use the same profiling scope.

Stereo fusion adds 3\,ms to the joint-attention control. The full pipeline is 310\,ms faster than FoundationStereo with 3DDA and 79\,ms faster than FastFoundationStereo with 3DDA. 

\subsection{Removing Wrist Observations}
Removing the wrist camera reduces StereoPolicy success from 96\% to 94\%, compared with 90\% to 86\% for RGB, 88\% to 84\% for RGB-D, and 76\% to 62\% for DP3 (Table~\ref{tab:nowrist}). StereoPolicy has a smaller decrease than these baselines and retains the highest absolute success rate, supporting the value of external stereo observations when wrist sensing is unavailable.
\input{tables/nowrist}

\begin{figure}[t]
    \centering
    \includegraphics[width=1\linewidth]{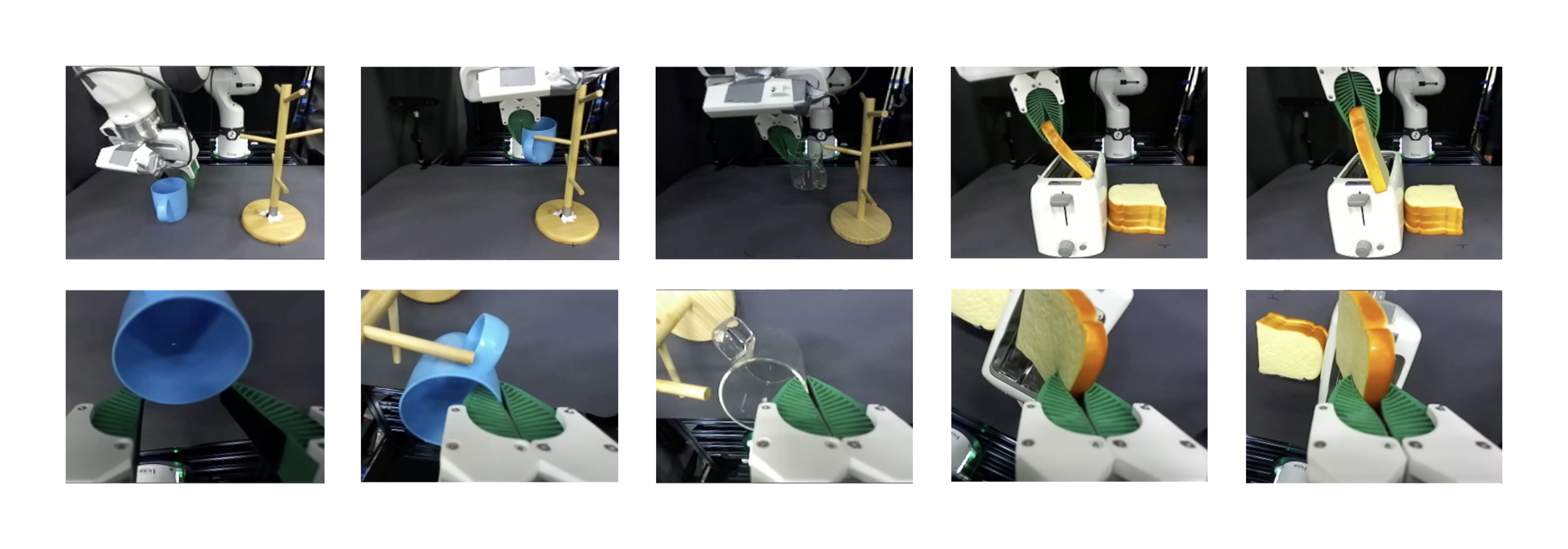}
    \caption{Failure cases of baseline visual modalities on real-world tabletop tasks. RGB, RGB-D, and point-cloud inputs can be inaccurate for fine-grained manipulation: gripper-target alignment can drift, reflective objects can obscure object boundaries, and transparent objects often produce missing or noisy depth and point-cloud geometry.}
    \label{fig:failure_tabletop}
\end{figure}

\begin{figure}[t]
    \centering
    \includegraphics[width=1\linewidth]{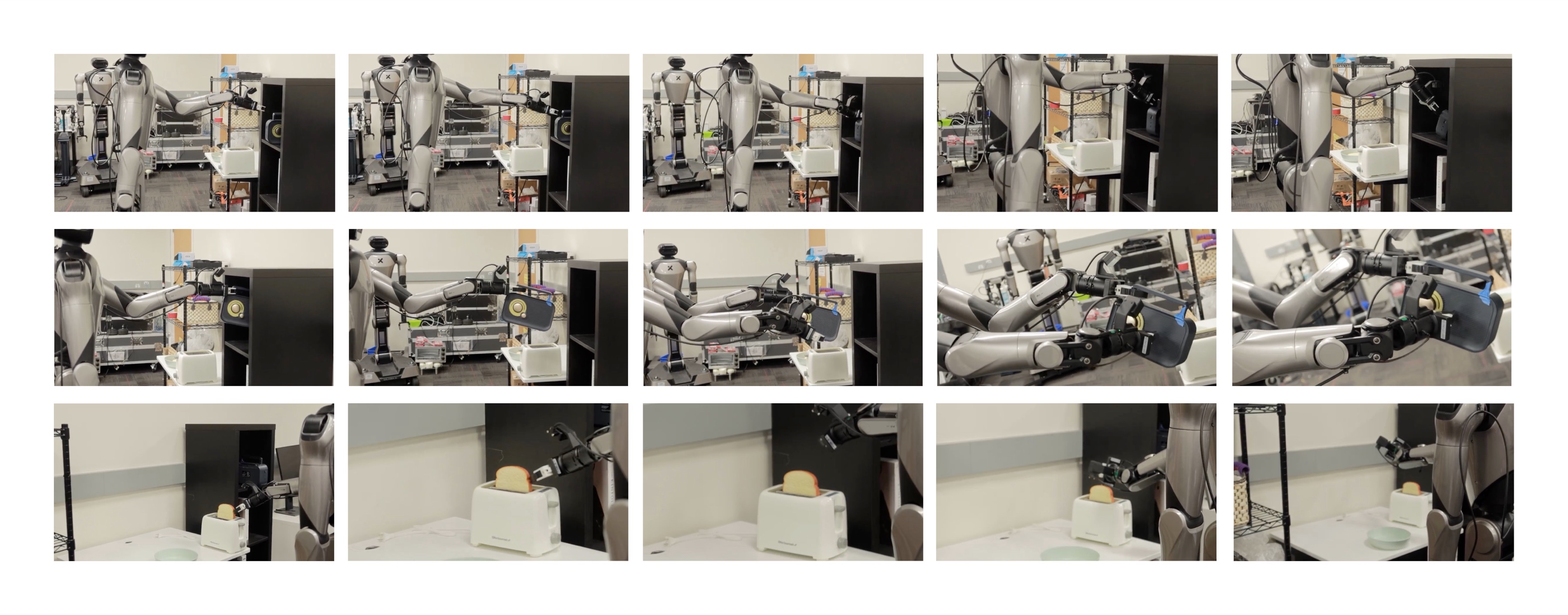}
    \caption{Monocular RGB failures in bimanual mobile manipulation. In the \textsc{Turn on Radio} task, RGB policies often fail at precise handle insertion and button pressing due to insufficient depth cues. In the \textsc{Toast PnP} task, failures commonly occur during grasp alignment and placement due to ambiguous monocular depth cues.}
    \label{fig:failure_mobile}
\end{figure}

\section{Qualitative Failure Cases}
\label{app:failure}

\paragraph{Baseline modality failures in tabletop manipulation.}
Figure~\ref{fig:failure_tabletop} shows representative tabletop failures caused by unreliable object geometry and spatial alignment. In Cup Hang tasks, weak object-boundary perception can lead to inaccurate cup localization and failed grasping. Noisy depth observations further cause the cup handle to misalign with the rack peg during hanging. For Glass Cup Hang tasks, specular reflection and transparency can make the cup partially or completely disappear in depth videos and reconstructed point clouds while it is lifted in free space. In Toast Insert tasks, insufficient 3D spatial information or inaccurate depth estimates can cause vertical or lateral misalignment between the toast and the toaster slot.

\paragraph{Remaining difficulty on transparent objects.}
StereoPolicy succeeds in only 3/20 Glass Cup Hang trials (Table~\ref{tab:real_result}), so transparent-object manipulation remains difficult despite its improvement over the tested baselines. Refraction and weak visual boundaries, especially around the thin glass rim and handle, can make localization and left/right correspondence ambiguous even in RGB images. Avoiding unreliable depth reconstruction therefore does not eliminate the visual ambiguity that can hinder grasping and handle--peg alignment.

\paragraph{Monocular RGB failures in mobile manipulation.}
Figure~\ref{fig:failure_mobile} visualizes representative failure cases of RGB-based VLA policies on mobile manipulation tasks. In the \textsc{Turn on Radio} task, the RGB policy often approaches the radio but fails to insert the gripper into the handle or press the button accurately. In the \textsc{Toast PnP} task, RGB policies can localize the toast but often fail during grasp alignment or final placement, where precise spatial alignment is required. In contrast, \algo uses stereo cues to improve gripper--target alignment.

%% file: tables/no_dino_sim.tex
\begin{table}[t]
\centering
\footnotesize
\begin{minipage}[t]{0.69\textwidth}
\vspace{0pt}
\centering
\setlength{\tabcolsep}{3.5pt}
\resizebox{\linewidth}{!}{%
\begin{tabular}{lcccc}
\toprule
\textbf{Benchmark} &
\multicolumn{2}{c}{\textsc{Robomimic}} &
\textsc{OmniGibson} & \\
\cmidrule(lr){2-3} \cmidrule(lr){4-4}
\textbf{Task} &
\textsc{ToolHang-100} &
\textsc{Square-100} &
\textsc{PourWater-200} &
Avg. SR \\
\midrule
\algodp & $94_{1.9}$ & $88_{0.9}$ & $70_{2.5}$ & \textbf{84.0} \\
\quad with (L,L) input & $62_{1.6}$ & $76_{1.6}$ & $50_{3.3}$ & 62.7\\
\quad w/o StereoTransformer & $82_{2.9}$ & $76_{2.8}$ & $52_{1.6}$ & 70.0\\
\quad w/o FPN & $78_{1.6}$ & $80_{1.6}$ & $64_{1.6}$ & 74.0 \\
\quad w/o DINO & $84_{1.6}$ & $82_{2.8}$ & $62_{2.8}$ & 76.0 \\
\quad with DINO on both view & $88_{1.6}$ & $84_{0.0}$ & $66_{3.3}$ & 79.3 \\
\bottomrule
\end{tabular}%
}
\caption{\textbf{Component Ablation of \algodp}. ToolHang and Square use 100 demonstrations; PourWater uses 200. Subscripts are standard deviations over three seeds.}
\label{tab:no_dino_sim}
\end{minipage}%
\hfill
\begin{minipage}[t]{0.28\textwidth}
\vspace{0pt}
\centering
\resizebox{\linewidth}{!}{%
\begin{tabular}{lr}
\toprule
Encoder & SR (\%) \\
\midrule
ResNet18 & $64_{2.8}$ \\
OpenCLIP-B/16 & $78_{2.8}$ \\
SigLIP-SO400M/14 & $80_{4.3}$ \\
EdgeNeXt-S & $82_{1.6}$ \\
ResNet18+FPN & $84_{1.6}$ \\
\bottomrule
\end{tabular}%
}
\caption{Vision-encoder comparison (w/o DINO) on ToolHang with 100 demonstrations. }
\label{tab:encoder_sweep}
\end{minipage}
\end{table}

%% file: tables/nowrist.tex
\begin{table}[!htb]
\centering

\resizebox{\linewidth}{!}{%
\begin{tabular}{lcccccccc}
\toprule
\textbf{Setting} 
& \multicolumn{1}{c}{\textbf{RGB}} 
& \multicolumn{1}{c}{\textbf{RGB-D}} 
& \multicolumn{3}{c}{\textbf{PCD}} 
& \multicolumn{1}{c}{\textbf{RGB-D}} 
& \multicolumn{1}{c}{\textbf{Multi-View}} 
& \multicolumn{1}{c}{\textbf{\algodp}} \\
\cmidrule(lr){4-6} \cmidrule(lr){7-7} 
&  
&  
& PointNet 
& PointNet++ 
& DP3 
& 3DDA 
&  
&  \\
\midrule
Default (w/ wrist camera) 
& $90_{1.6}$ 
& $88_{2.5}$ 
& $32_{2.8}$ 
& $28_{4.9}$ 
& $76_{0.9}$ 
& $92_{2.8}$ 
& $92_{1.9}$ 
& $\mathbf{96}_{2.8}$ \\

w/o wrist camera 
& $86_{1.6}$ 
& $84_{1.6}$ 
& $30_{4.9}$ 
& $28_{3.3}$ 
& $62_{2.8}$ 
& $88_{1.6}$ 
& $88_{1.6}$ 
& $\mathbf{94}_{3.3}$ \\
\bottomrule
\end{tabular}
}

\caption{Effect of removing wrist-camera observations on ToolHang with 200 demos. Success rate (\%); subscripts are standard deviations over three seeds.}
\label{tab:nowrist}
\end{table}

%% file: main.bib
@String(CVPR= {IEEE Conf. Comput. Vis. Pattern Recog.})

@String(ICCV= {Int. Conf. Comput. Vis.})

@String(ECCV= {Eur. Conf. Comput. Vis.})

@String(ICLR = {Int. Conf. Learn. Represent.})

@String(CVPR  = {CVPR})

@String(ICCV  = {ICCV})

@String(ECCV  = {ECCV})

@String(ICLR  = {ICLR})

@inproceedings{robocasa,
  title     = {{RoboCasa}: Large-Scale Simulation of Household Tasks for Generalist Robots},
  author    = {Soroush Nasiriany and Abhiram Maddukuri and Lance Zhang and Adeet Parikh and Aaron Lo and Abhishek Joshi and Ajay Mandlekar and Yuke Zhu},
  booktitle = {Robotics: Science and Systems (RSS)},
  year      = {2024}
}

@inproceedings{robomimic,
  title     = {What Matters in Learning from Offline Human Demonstrations for Robot Manipulation},
  author    = {Ajay Mandlekar and Danfei Xu and Josiah Wong and Soroush Nasiriany and Chen Wang and Rohun Kulkarni and Li Fei-Fei and Silvio Savarese and Yuke Zhu and Roberto Mart{\'i}n-Mart{\'i}n},
  booktitle = {Proceedings of the 5th Conference on Robot Learning (CoRL)},
  volume    = {164},
  series    = {Proceedings of Machine Learning Research},
  pages     = {1678--1690},
  publisher = {PMLR},
  year      = {2021}
}

@article{behavior,
    title   = {{BEHAVIOR-1K}: A Human-Centered, Embodied {AI} Benchmark with 1,000 Everyday Activities and Realistic Simulation},
    author  = {Chengshu Li and Ruohan Zhang and Josiah Wong and Cem Gokmen and Sanjana Srivastava and Roberto Martín-Martín and Chen Wang and Gabrael Levine and Wensi Ai and Benjamin Martinez and Hang Yin and Michael Lingelbach and Minjune Hwang and Ayano Hiranaka and Sujay Garlanka and Arman Aydin and Sharon Lee and Jiankai Sun and Mona Anvari and Manasi Sharma and Dhruva Bansal and Samuel Hunter and Kyu-Young Kim and Alan Lou and Caleb R Matthews and Ivan Villa-Renteria and Jerry Huayang Tang and Claire Tang and Fei Xia and Yunzhu Li and Silvio Savarese and Hyowon Gweon and C. Karen Liu and Jiajun Wu and Li Fei-Fei},
    journal = {arXiv preprint arXiv:2403.09227},
    year    = {2024}
}

@inproceedings{radford2021learningtransferablevisualmodels,
  title     = {Learning Transferable Visual Models From Natural Language Supervision},
  author    = {Alec Radford and Jong Wook Kim and Chris Hallacy and Aditya Ramesh and Gabriel Goh and Sandhini Agarwal and Girish Sastry and Amanda Askell and Pamela Mishkin and Jack Clark and Gretchen Krueger and Ilya Sutskever},
  booktitle = {Proceedings of the 38th International Conference on Machine Learning (ICML)},
  series    = {Proceedings of Machine Learning Research},
  volume    = {139},
  pages     = {8748--8763},
  year      = {2021},
  url       = {https://proceedings.mlr.press/v139/radford21a.html}
}

@inproceedings{dosovitskiy2021imageworth16x16words,
  title     = {An Image is Worth 16x16 Words: Transformers for Image Recognition at Scale},
  author    = {Alexey Dosovitskiy and Lucas Beyer and Alexander Kolesnikov and Dirk Weissenborn and Xiaohua Zhai and Thomas Unterthiner and Mostafa Dehghani and Matthias Minderer and Georg Heigold and Sylvain Gelly and Jakob Uszkoreit and Neil Houlsby},
  booktitle = {International Conference on Learning Representations (ICLR)},
  year      = {2021},
  url       = {https://openreview.net/forum?id=YicbFdNTTy}
}

@inproceedings{wu2024pointtransformerv3simpler,
  title     = {{Point Transformer V3}: Simpler, Faster, Stronger},
  author    = {Xiaoyang Wu and Li Jiang and Peng-Shuai Wang and Zhijian Liu and Xihui Liu and Yu Qiao and Wanli Ouyang and Tong He and Hengshuang Zhao},
  booktitle = {Proceedings of the IEEE/CVF Conference on Computer Vision and Pattern Recognition (CVPR)},
  pages     = {4840--4851},
  year      = {2024},
  doi       = {10.1109/CVPR52733.2024.00463}
}

@inproceedings{wen2025foundationstereozeroshotstereomatching,
  title     = {{FoundationStereo}: Zero-Shot Stereo Matching},
  author    = {Bowen Wen and Matthew Trepte and Joseph Aribido and Jan Kautz and Orazio Gallo and Stan Birchfield},
  booktitle = {Proceedings of the IEEE/CVF Conference on Computer Vision and Pattern Recognition (CVPR)},
  pages     = {5249--5260},
  year      = {2025},
  doi       = {10.1109/CVPR52734.2025.00495}
}

@inproceedings{min2025stextsuperscript2mtextsuperscript2scalablestereomatching,
  title     = {{S\textsuperscript{2}M\textsuperscript{2}}: Scalable Stereo Matching Model for Reliable Depth Estimation},
  author    = {Junhong Min and Youngpil Jeon and Jimin Kim and Minyong Choi},
  booktitle = {Proceedings of the IEEE/CVF International Conference on Computer Vision (ICCV)},
  pages     = {26729--26739},
  year      = {2025},
  url       = {https://arxiv.org/abs/2507.13229}
}

@inproceedings{lipson2021raftstereomultilevelrecurrentfield,
  title     = {{RAFT-Stereo}: Multilevel Recurrent Field Transforms for Stereo Matching},
  author    = {Lahav Lipson and Zachary Teed and Jia Deng},
  booktitle = {International Conference on 3D Vision (3DV)},
  pages     = {218--227},
  year      = {2021},
  doi       = {10.1109/3DV53792.2021.00032},
  url       = {https://arxiv.org/abs/2109.07547}
}

@inproceedings{bartolomei2025stereoanywhererobustzeroshot,
  title     = {{Stereo Anywhere}: Robust Zero-Shot Deep Stereo Matching Even Where Either Stereo or Mono Fail},
  author    = {Luca Bartolomei and Fabio Tosi and Matteo Poggi and Stefano Mattoccia},
  booktitle = {Proceedings of the IEEE/CVF Conference on Computer Vision and Pattern Recognition (CVPR)},
  pages     = {1013--1027},
  year      = {2025},
  doi       = {10.1109/CVPR52734.2025.00103},
  url       = {https://arxiv.org/abs/2412.04472}
}

@article{xu2025igeviterativemultirangegeometry,
  title     = {{IGEV++}: Iterative Multi-Range Geometry Encoding Volumes for Stereo Matching},
  author    = {Gangwei Xu and Xianqi Wang and Zhaoxing Zhang and Junda Cheng and Chunyuan Liao and Xin Yang},
  journal   = {IEEE Transactions on Pattern Analysis and Machine Intelligence},
  volume    = {47},
  number    = {8},
  pages     = {7108--7122},
  year      = {2025},
  doi       = {10.1109/TPAMI.2025.3569218}
}

@inproceedings{chang2018pyramidstereomatchingnetwork,
  title     = {Pyramid Stereo Matching Network},
  author    = {Jia-Ren Chang and Yong-Sheng Chen},
  booktitle = {Proceedings of the IEEE/CVF Conference on Computer Vision and Pattern Recognition (CVPR)},
  pages     = {5410--5418},
  year      = {2018},
  doi       = {10.1109/CVPR.2018.00567}
}

@inproceedings{wang2025robustereorobustzeroshotstereo,
  title     = {{RobuSTereo}: Robust Zero-Shot Stereo Matching under Adverse Weather},
  author    = {Yuran Wang and Yingping Liang and Yutao Hu and Ying Fu},
  booktitle = {Proceedings of the IEEE/CVF International Conference on Computer Vision (ICCV)},
  pages     = {25134--25144},
  year      = {2025}
}

@inproceedings{singh2024dextrah,
  title={{DextrAH-RGB}: Visuomotor Policies to Grasp Anything with Dexterous Hands},
  author={Ritvik Singh and Arthur Allshire and Ankur Handa and Nathan Ratliff and Karl Van Wyk},
  booktitle={Proceedings of the IEEE International Conference on Robotics and Automation (ICRA)},
  year={2026}
}

@inproceedings{lum2024dextrah,
  title     = {{DextrAH-G}: Pixels-to-Action Dexterous Arm-Hand Grasping with Geometric Fabrics},
  author    = {Lum, Tyler Ga Wei and Matak, Martin and Makoviychuk, Viktor and Handa, Ankur and Allshire, Arthur and Hermans, Tucker and Ratliff, Nathan D. and Van Wyk, Karl},
  booktitle = {Proceedings of The 8th Conference on Robot Learning (CoRL)},
  volume    = {270},
  pages     = {3182--3211},
  publisher = {PMLR},
  year      = {2024},
  url       = {https://arxiv.org/abs/2407.02274}
}

@inproceedings{kollar2021simnetenablingrobustunknown,
  title     = {{SimNet}: Enabling Robust Unknown Object Manipulation from Pure Synthetic Data via Stereo},
  author    = {Thomas Kollar and Michael Laskey and Kevin Stone and Brijen Thananjeyan and Mark Tjersland},
  booktitle = {Proceedings of the 5th Conference on Robot Learning (CoRL)},
  series    = {Proceedings of Machine Learning Research},
  volume    = {164},
  pages     = {938--948},
  publisher = {PMLR},
  year      = {2021},
  url       = {https://proceedings.mlr.press/v164/kollar22a.html}
}

@article{shankar2021learnedstereodepthrobotic,
  title={A Learned Stereo Depth System for Robotic Manipulation in Homes},
  author={Krishna Shankar and Mark Tjersland and Jeremy Ma and Kevin Stone and Max Bajracharya},
  journal={IEEE Robotics and Automation Letters},
  volume={7},
  number={2},
  pages={2305--2312},
  year={2022},
  doi={10.1109/LRA.2022.3143895}
}

@article{bai2025cleardepthenhancedstereoperception,
  title={{ClearDepth}: Enhanced Stereo Perception of Transparent Objects for Robotic Manipulation},
  author={Kaixin Bai and Huajian Zeng and Lei Zhang and Yiwen Liu and Hongli Xu and Zhaopeng Chen and Jianwei Zhang},
  journal={arXiv preprint arXiv:2409.08926},
  year={2024},
  url={https://arxiv.org/abs/2409.08926}
}

@article{dp,
  author  = {Cheng Chi and Zhenjia Xu and Siyuan Feng and Eric Cousineau and Yilun Du and Benjamin Burchfiel and Russ Tedrake and Shuran Song},
  title   = {{Diffusion Policy}: Visuomotor Policy Learning via Action Diffusion},
  journal = {The International Journal of Robotics Research},
  volume  = {44},
  number  = {10-11},
  pages   = {1684--1704},
  year    = {2024},
  doi     = {10.1177/02783649241273668}
}

@inproceedings{resnet,
  title     = {Deep Residual Learning for Image Recognition},
  author    = {Kaiming He and Xiangyu Zhang and Shaoqing Ren and Jian Sun},
  booktitle = {Proceedings of the IEEE Conference on Computer Vision and Pattern Recognition (CVPR)},
  pages     = {770--778},
  year      = {2016}
}

@inproceedings{wu2018groupnormalization,
  title     = {Group Normalization},
  author    = {Yuxin Wu and Kaiming He},
  booktitle = {Proceedings of the European Conference on Computer Vision (ECCV)},
  pages     = {3--19},
  year      = {2018},
  doi       = {10.1007/978-3-030-01261-8_1}
}

@misc{curobo_report23,
      title={{cuRobo}: Parallelized Collision-Free Minimum-Jerk Robot Motion Generation},
      author={Balakumar Sundaralingam and Siva Kumar Sastry Hari and Adam Fishman and Caelan Garrett
              and Karl Van Wyk and Valts Blukis and Alexander Millane and Helen Oleynikova and Ankur Handa
              and Fabio Ramos and Nathan Ratliff and Dieter Fox},
      year={2023},
      eprint={2310.17274},
      archivePrefix={arXiv},
      primaryClass={cs.RO}
}

@article{gr00tn1_2025,
  title   = {{GR00T} {N1}: An Open Foundation Model for Generalist Humanoid Robots},
  author  = {{NVIDIA} and Johan Bjorck and Fernando Casta{\~n}eda and Nikita Cherniadev and Xingye Da and Runyu Ding and Linxi Fan and Yu Fang and Dieter Fox and Fengyuan Hu and Spencer Huang and Joel Jang and Zhenyu Jiang and Jan Kautz and Kaushil Kundalia and Lawrence Lao and Zhiqi Li and Zongyu Lin and Kevin Lin and Guilin Liu and Edith Llontop and Loic Magne and Ajay Mandlekar and Avnish Narayan and Soroush Nasiriany and Scott Reed and You Liang Tan and Guanzhi Wang and Zu Wang and Jing Wang and Qi Wang and Jiannan Xiang and Yuqi Xie and Yinzhen Xu and Zhenjia Xu and Seonghyeon Ye and Zhiding Yu and Ao Zhang and Hao Zhang and Yizhou Zhao and Ruijie Zheng and Yuke Zhu},
  journal = {arXiv preprint arXiv:2503.14734},
  year    = {2025},
  url     = {https://arxiv.org/abs/2503.14734}
}

@inproceedings{peri2024pointcloudmodelsimprove,
  title     = {Point Cloud Models Improve Visual Robustness in Robotic Learners},
  author    = {Skand Peri and Iain Lee and Chanho Kim and Li Fuxin and Tucker Hermans and Stefan Lee},
  booktitle = {IEEE International Conference on Robotics and Automation (ICRA)},
  pages     = {17529--17536},
  year      = {2024},
  doi       = {10.1109/ICRA57147.2024.10610710},
  url       = {https://arxiv.org/abs/2404.18926}
}

@article{levine2016endtoendtrainingdeepvisuomotor,
  title   = {End-to-End Training of Deep Visuomotor Policies},
  author  = {Sergey Levine and Chelsea Finn and Trevor Darrell and Pieter Abbeel},
  journal = {Journal of Machine Learning Research},
  volume  = {17},
  number  = {39},
  pages   = {1--40},
  year    = {2016},
  url     = {https://jmlr.org/papers/v17/15-522.html}
}

@inproceedings{zhu2018reinforcementimitationlearningdiverse,
  title     = {Reinforcement and Imitation Learning for Diverse Visuomotor Skills},
  author    = {Yuke Zhu and Ziyu Wang and Josh Merel and Andrei Rusu and Tom Erez and Serkan Cabi and Saran Tunyasuvunakool and J{\'a}nos Kram{\'a}r and Raia Hadsell and Nando de Freitas and Nicolas Heess},
  booktitle = {Proceedings of Robotics: Science and Systems (RSS)},
  year      = {2018},
  doi       = {10.15607/RSS.2018.XIV.009},
  url       = {https://www.roboticsproceedings.org/rss14/p09.html}
}

@inproceedings{brohan2022rt1,
  title     = {{RT-1}: Robotics Transformer for Real-World Control at Scale},
  author    = {Anthony Brohan and Noah Brown and Justice Carbajal and Yevgen Chebotar and Joseph Dabis and Chelsea Finn and Keerthana Gopalakrishnan and Karol Hausman and Alexander Herzog and Jasmine Hsu and Julian Ibarz and Brian Ichter and Alex Irpan and Tomas Jackson and Sally Jesmonth and Nikhil Joshi and Ryan Julian and Dmitry Kalashnikov and Yuheng Kuang and Isabel Leal and Kuang-Huei Lee and Sergey Levine and Yao Lu and Utsav Malla and Deeksha Manjunath and Igor Mordatch and Ofir Nachum and Carolina Parada and Jodilyn Peralta and Emily Perez and Karl Pertsch and Jornell Quiambao and Kanishka Rao and Michael S. Ryoo and Grecia Salazar and Pannag R. Sanketi and Kevin Sayed and Jaspiar Singh and Sumedh Sontakke and Austin Stone and Clayton Tan and Huong Tran and Vincent Vanhoucke and Steve Vega and Quan H. Vuong and Fei Xia and Ted Xiao and Peng Xu and Sichun Xu and Tianhe Yu and Brianna Zitkovich},
  booktitle = {Proceedings of Robotics: Science and Systems (RSS)},
  year      = {2023},
  doi       = {10.15607/RSS.2023.XIX.025},
  url       = {https://arxiv.org/abs/2212.06817}
}

@article{reed2022gato,
  title   = {A Generalist Agent},
  author  = {Scott Reed and Konrad Zolna and Emilio Parisotto and Sergio G{\'o}mez Colmenarejo and Alexander Novikov and Gabriel Barth-Maron and Mai Gim{\'e}nez and Yury Sulsky and Jackie Kay and Jost Tobias Springenberg and Tom Eccles and Jake Bruce and Ali Razavi and Ashley Edwards and Nicolas Heess and Yutian Chen and Raia Hadsell and Oriol Vinyals and Mahyar Bordbar and Nando de Freitas},
  journal = {Transactions on Machine Learning Research},
  year    = {2022},
  url     = {https://openreview.net/forum?id=1ikK0kHjvj}
}

@inproceedings{brohan2023rt2,
  title     = {{RT-2}: Vision-Language-Action Models Transfer Web Knowledge to Robotic Control},
  author    = {Anthony Brohan and Noah Brown and Justice Carbajal and Yevgen Chebotar and Xi Chen and Krzysztof Marcin Choromanski and Tianli Ding and Danny Driess and Kumar Avinava Dubey and Chelsea Finn and Pete Florence and Chuyuan Fu and Montserrat {Gonzalez Arenas} and Keerthana Gopalakrishnan and Kehang Han and Karol Hausman and Alexander Herzog and Jasmine Hsu and Brian Ichter and Alex Irpan and Nikhil J. Joshi and Ryan Julian and Dmitry Kalashnikov and Yuheng Kuang and Isabel Leal and Lisa Lee and Tsang-Wei Edward Lee and Sergey Levine and Yao Lu and Henryk Michalewski and Igor Mordatch and Karl Pertsch and Kanishka Rao and Krista Reymann and Michael S. Ryoo and Grecia Salazar and Pannag R. Sanketi and Pierre Sermanet and Jaspiar Singh and Anikait Singh and Radu Soricut and Huong Tran and Vincent Vanhoucke and Quan Vuong and Ayzaan Wahid and Stefan Welker and Paul Wohlhart and Jialin Wu and Fei Xia and Ted Xiao and Peng Xu and Sichun Xu and Tianhe Yu and Brianna Zitkovich},
  booktitle = {Proceedings of the 7th Conference on Robot Learning (CoRL)},
  series    = {Proceedings of Machine Learning Research},
  volume    = {229},
  pages     = {2165--2183},
  publisher = {PMLR},
  year      = {2023},
  url       = {https://proceedings.mlr.press/v229/zitkovich23a.html}
}

@inproceedings{
kim2024openvla,
title={Open{VLA}: An Open-Source Vision-Language-Action Model},
author={Moo Jin Kim and Karl Pertsch and Siddharth Karamcheti and Ted Xiao and Ashwin Balakrishna and Suraj Nair and Rafael Rafailov and Ethan P Foster and Pannag R Sanketi and Quan Vuong and Thomas Kollar and Benjamin Burchfiel and Russ Tedrake and Dorsa Sadigh and Sergey Levine and Percy Liang and Chelsea Finn},
booktitle={8th Annual Conference on Robot Learning},
year={2024},
url={https://openreview.net/forum?id=ZMnD6QZAE6}
}

@inproceedings{aloha,
  author       = {Tony Z. Zhao and
                  Vikash Kumar and
                  Sergey Levine and
                  Chelsea Finn},
  editor       = {Kostas E. Bekris and
                  Kris Hauser and
                  Sylvia L. Herbert and
                  Jingjin Yu},
  title        = {Learning Fine-Grained Bimanual Manipulation with Low-Cost Hardware},
  booktitle    = {Robotics: Science and Systems XIX, Daegu, Republic of Korea, July
                  10-14, 2023},
  year         = {2023},
  url          = {https://doi.org/10.15607/RSS.2023.XIX.016},
  doi          = {10.15607/RSS.2023.XIX.016},
}

@inproceedings{black2024_0,
  title     = {$\pi_0$: A Vision-Language-Action Flow Model for General Robot Control},
  author    = {Kevin Black and Noah Brown and Danny Driess and Adnan Esmail and Michael Robert Equi and Chelsea Finn and Niccolo Fusai and Lachy Groom and Karol Hausman and Brian Ichter and Szymon Jakubczak and Tim Jones and Liyiming Ke and Sergey Levine and Adrian Li-Bell and Mohith Mothukuri and Suraj Nair and Karl Pertsch and Lucy Xiaoyang Shi and Laura Smith and James Tanner and Quan Vuong and Anna Walling and Haohuan Wang and Ury Zhilinsky},
  booktitle = {Proceedings of Robotics: Science and Systems (RSS)},
  year      = {2025},
  doi       = {10.15607/RSS.2025.XXI.010},
  url       = {https://arxiv.org/abs/2410.24164}
}

@inproceedings{shridhar2021cliport,
  title={{CLIPort}: What and Where Pathways for Robotic Manipulation},
  author={Mohit Shridhar and Lucas Manuelli and Dieter Fox},
  booktitle={Proceedings of the 5th Conference on Robot Learning (CoRL)},
  series={Proceedings of Machine Learning Research},
  volume={164},
  pages={894--906},
  publisher={PMLR},
  year={2021}
}

@inproceedings{jiang2022vima,
  title     = {{VIMA}: Robot Manipulation with Multimodal Prompts},
  author    = {Yunfan Jiang and Agrim Gupta and Zichen Zhang and Guanzhi Wang and Yongqiang Dou and Yanjun Chen and Li Fei-Fei and Anima Anandkumar and Yuke Zhu and Linxi Fan},
  booktitle = {Proceedings of the 40th International Conference on Machine Learning},
  series    = {Proceedings of Machine Learning Research},
  volume    = {202},
  pages     = {14975--15022},
  publisher = {PMLR},
  year      = {2023},
  url       = {https://proceedings.mlr.press/v202/jiang23b.html}
}

@inproceedings{liu2024visual,
  title     = {Visual Whole-Body Control for Legged Loco-Manipulation},
  author    = {Minghuan Liu and Zixuan Chen and Xuxin Cheng and Yandong Ji and Ri-Zhao Qiu and Ruihan Yang and Xiaolong Wang},
  booktitle = {Proceedings of The 8th Conference on Robot Learning (CoRL)},
  volume    = {270},
  pages     = {234--257},
  publisher = {PMLR},
  year      = {2024},
  url       = {https://arxiv.org/abs/2403.16967}
}

@inproceedings{uppal2024spin,
  author    = {Shagun Uppal and Ananye Agarwal and Haoyu Xiong and Kenneth Shaw and Deepak Pathak},
  title     = {{SPIN}: Simultaneous Perception, Interaction and Navigation},
  booktitle = {Proceedings of the IEEE/CVF Conference on Computer Vision and Pattern Recognition (CVPR)},
  pages     = {18133--18142},
  year      = {2024}
}

@inproceedings{jiang2025brs,
title={{BEHAVIOR} Robot Suite: Streamlining Real-World Whole-Body Manipulation for Everyday Household Activities},
author={Yunfan Jiang and Ruohan Zhang and Josiah Wong and Chen Wang and Yanjie Ze and Hang Yin and Cem Gokmen and Shuran Song and Jiajun Wu and Li Fei-Fei},
booktitle={9th Annual Conference on Robot Learning},
year={2025},
url={https://openreview.net/forum?id=v2KevjWScT}
}

@inproceedings{sundaresan2025homerlearninginthewildmobile,
  title     = {{HoMeR}: Learning In-the-Wild Mobile Manipulation via Hybrid Imitation and Whole-Body Control},
  author    = {Priya Sundaresan and Rhea Malhotra and Phillip Miao and Jingyun Yang and Jimmy Wu and Hengyuan Hu and Rika Antonova and Francis Engelmann and Dorsa Sadigh and Jeannette Bohg},
  booktitle = {IEEE International Conference on Robotics and Automation (ICRA)},
  year      = {2026},
  note      = {arXiv:2506.01185}
}

@inproceedings{yang2024equibotsim3equivariantdiffusionpolicy,
  title     = {{EquiBot}: {SIM(3)}-Equivariant Diffusion Policy for Generalizable and Data Efficient Learning},
  author    = {Jingyun Yang and Zi-ang Cao and Congyue Deng and Rika Antonova and Shuran Song and Jeannette Bohg},
  booktitle = {Proceedings of the 8th Conference on Robot Learning (CoRL)},
  year      = {2024},
  volume    = {270},
  pages     = {1048--1068},
  publisher = {PMLR},
  url       = {https://proceedings.mlr.press/v270/yang25a.html}
}

@inproceedings{qin2022dexpoint0,
  title     = {{DexPoint}: Generalizable Point Cloud Reinforcement Learning for Sim-to-Real Dexterous Manipulation},
  author    = {Yuzhe Qin and Binghao Huang and Zhao-Heng Yin and Hao Su and Xiaolong Wang},
  booktitle = {Conference on Robot Learning (CoRL)},
  series    = {Proceedings of Machine Learning Research},
  volume    = {205},
  pages     = {594--605},
  year      = {2022},
  url       = {https://proceedings.mlr.press/v205/qin23a.html}
}

@inproceedings{jiang2024transic,
  title     = {{TRANSIC}: Sim-to-Real Policy Transfer by Learning from Online Correction},
  author    = {Yunfan Jiang and Chen Wang and Ruohan Zhang and Jiajun Wu and Li Fei-Fei},
  booktitle = {Proceedings of The 8th Conference on Robot Learning},
  series    = {Proceedings of Machine Learning Research},
  volume    = {270},
  pages     = {1691--1729},
  publisher = {PMLR},
  year      = {2024},
  url       = {https://proceedings.mlr.press/v270/jiang25a.html}
}

@inproceedings{wang2024dexcap,
  title     = {{DexCap}: Scalable and Portable Mocap Data Collection System for Dexterous Manipulation},
  author    = {Chen Wang and Haochen Shi and Weizhuo Wang and Ruohan Zhang and Fei-Fei Li and Karen Liu},
  booktitle = {Proceedings of Robotics: Science and Systems (RSS)},
  year      = {2024},
  doi       = {10.15607/RSS.2024.XX.043}
}

@article{
doi:10.1126/science.968482,
author = {D. Marr  and T. Poggio },
title = {Cooperative Computation of Stereo Disparity},
journal = {Science},
volume = {194},
number = {4262},
pages = {283-287},
year = {1976},
doi = {10.1126/science.968482},
URL = {https://www.science.org/doi/abs/10.1126/science.968482},
eprint = {https://www.science.org/doi/pdf/10.1126/science.968482}}

@inproceedings{Shen2021CFNetCA,
  title     = {{CFNet}: Cascade and Fused Cost Volume for Robust Stereo Matching},
  author    = {Zhelun Shen and Yuchao Dai and Zhibo Rao},
  booktitle = {Proceedings of the IEEE/CVF Conference on Computer Vision and Pattern Recognition (CVPR)},
  year      = {2021},
  pages     = {13906--13915},
  url       = {https://openaccess.thecvf.com/content/CVPR2021/html/Shen_CFNet_Cascade_and_Fused_Cost_Volume_for_Robust_Stereo_Matching_CVPR_2021_paper.html}
}

@inproceedings{Xu2020AANetAA,
  title={{AANet}: Adaptive Aggregation Network for Efficient Stereo Matching},
  author={Haofei Xu and Juyong Zhang},
  booktitle={Proceedings of the IEEE/CVF Conference on Computer Vision and Pattern Recognition (CVPR)},
  year={2020},
  pages={1956--1965},
  doi={10.1109/CVPR42600.2020.00203}
}

@InProceedings{Li_2022_CVPR,
    author    = {Li, Jiankun and Wang, Peisen and Xiong, Pengfei and Cai, Tao and Yan, Ziwei and Yang, Lei and Liu, Jiangyu and Fan, Haoqiang and Liu, Shuaicheng},
    title     = {Practical Stereo Matching via Cascaded Recurrent Network With Adaptive Correlation},
    booktitle = {Proceedings of the IEEE/CVF Conference on Computer Vision and Pattern Recognition (CVPR)},
    month     = {June},
    year      = {2022},
    pages     = {16263-16272}
}

@InProceedings{Poggi_2020_CVPR,
author = {Poggi, Matteo and Aleotti, Filippo and Tosi, Fabio and Mattoccia, Stefano},
title = {On the Uncertainty of Self-Supervised Monocular Depth Estimation},
booktitle = {Proceedings of the IEEE/CVF Conference on Computer Vision and Pattern Recognition (CVPR)},
month = {June},
year = {2020}
}

@InProceedings{10.1007/978-3-030-58536-5_25,
author="Zhang, Feihu
and Qi, Xiaojuan
and Yang, Ruigang
and Prisacariu, Victor
and Wah, Benjamin
and Torr, Philip",
editor="Vedaldi, Andrea
and Bischof, Horst
and Brox, Thomas
and Frahm, Jan-Michael",
title="Domain-Invariant Stereo Matching Networks",
booktitle="Computer Vision -- ECCV 2020",
year="2020",
publisher="Springer International Publishing",
address="Cham",
pages="420--439",
isbn="978-3-030-58536-5"
}

@inproceedings{Weinzaepfel_2023_ICCV,
  author={Philippe Weinzaepfel and Thomas Lucas and Vincent Leroy and Yohann Cabon and Vaibhav Arora and Romain Br{\'e}gier and Gabriela Csurka and Leonid Antsfeld and Boris Chidlovskii and J{\'e}r{\^o}me Revaud},
  title={{CroCo v2}: Improved Cross-view Completion Pre-training for Stereo Matching and Optical Flow},
  booktitle={Proceedings of the IEEE/CVF International Conference on Computer Vision (ICCV)},
  year={2023},
  pages={17969--17980}
}

@InProceedings{Li_2021_ICCV,
    author    = {Li, Zhaoshuo and Liu, Xingtong and Drenkow, Nathan and Ding, Andy and Creighton, Francis X. and Taylor, Russell H. and Unberath, Mathias},
    title     = {Revisiting Stereo Depth Estimation From a Sequence-to-Sequence Perspective With Transformers},
    booktitle = {Proceedings of the IEEE/CVF International Conference on Computer Vision (ICCV)},
    month     = {October},
    year      = {2021},
    pages     = {6197-6206}
}

@InProceedings{Xu_2023_CVPR,
    author    = {Xu, Gangwei and Wang, Xianqi and Ding, Xiaohuan and Yang, Xin},
    title     = {Iterative Geometry Encoding Volume for Stereo Matching},
    booktitle = {Proceedings of the IEEE/CVF Conference on Computer Vision and Pattern Recognition (CVPR)},
    month     = {June},
    year      = {2023},
    pages     = {21919-21928}
}

@inproceedings{Yang2023NeuralVM,
  title={Neural Volumetric Memory for Visual Locomotion Control},
  author={Ruihan Yang and Ge Yang and Xiaolong Wang},
  booktitle={Proceedings of the IEEE/CVF Conference on Computer Vision and Pattern Recognition (CVPR)},
  year={2023},
  pages={1430--1440},
  doi={10.1109/CVPR52729.2023.00144}
}

@inproceedings{ze20243d,
  title     = {{3D Diffusion Policy}: Generalizable Visuomotor Policy Learning via Simple {3D} Representations},
  author    = {Yanjie Ze and Gu Zhang and Kangning Zhang and Chenyuan Hu and Muhan Wang and Huazhe Xu},
  booktitle = {Proceedings of Robotics: Science and Systems (RSS)},
  year      = {2024},
  doi       = {10.15607/RSS.2024.XX.067},
  url       = {https://www.roboticsproceedings.org/rss20/p067.html}
}

@inproceedings{Goyal2023RVTRV,
  title={{RVT}: Robotic View Transformer for {3D} Object Manipulation},
  author={Ankit Goyal and Jie Xu and Yijie Guo and Valts Blukis and Yu-Wei Chao and Dieter Fox},
  booktitle={Conference on Robot Learning (CoRL)},
  year={2023},
  pages={694--710},
  url={https://proceedings.mlr.press/v229/goyal23a.html}
}

@INPROCEEDINGS{6248074,
  author={Geiger, Andreas and Lenz, Philip and Urtasun, Raquel},
  booktitle={2012 IEEE Conference on Computer Vision and Pattern Recognition}, 
  title={Are we ready for autonomous driving? The {KITTI} vision benchmark suite}, 
  year={2012},
  volume={},
  number={},
  pages={3354-3361},
  doi={10.1109/CVPR.2012.6248074}}

@InProceedings{Menze_2015_CVPR,
author = {Menze, Moritz and Geiger, Andreas},
title = {Object Scene Flow for Autonomous Vehicles},
booktitle = {Proceedings of the IEEE Conference on Computer Vision and Pattern Recognition (CVPR)},
month = {June},
year = {2015}
}

@InProceedings{10.1007/978-3-319-11752-2_3,
author="Scharstein, Daniel
and Hirschm{\"u}ller, Heiko
and Kitajima, York
and Krathwohl, Greg
and Ne{\v{s}}i{\'{c}}, Nera
and Wang, Xi
and Westling, Porter",
editor="Jiang, Xiaoyi
and Hornegger, Joachim
and Koch, Reinhard",
title="High-Resolution Stereo Datasets with Subpixel-Accurate Ground Truth",
booktitle="Pattern Recognition",
year="2014",
publisher="Springer International Publishing",
address="Cham",
pages="31--42",
isbn="978-3-319-11752-2"
}

@article{qi2017pointnet,
  title   = {{PointNet++}: Deep hierarchical feature learning on point sets in a metric space},
  author  = {Qi, Charles Ruizhongtai and Yi, Li and Su, Hao and Guibas, Leonidas J},
  journal = {Advances in neural information processing systems},
  volume  = {30},
  year    = {2017}
}

@inproceedings{jaegle2021perceiver,
  title     = {Perceiver: General Perception with Iterative Attention},
  author    = {Andrew Jaegle and Felix Gimeno and Andy Brock and Oriol Vinyals and Andrew Zisserman and Joao Carreira},
  booktitle = {Proceedings of the 38th International Conference on Machine Learning},
  series    = {Proceedings of Machine Learning Research},
  volume    = {139},
  pages     = {4651--4664},
  publisher = {PMLR},
  year      = {2021},
  url       = {https://proceedings.mlr.press/v139/jaegle21a.html}
}

@inproceedings{Zhao2020PointT,
  title={{Point Transformer}},
  author={Hengshuang Zhao and Li Jiang and Jiaya Jia and Philip H. S. Torr and Vladlen Koltun},
  booktitle={Proceedings of the IEEE/CVF International Conference on Computer Vision (ICCV)},
  year={2021},
  pages={16239--16248},
  doi={10.1109/ICCV48922.2021.01595}
}

@InProceedings{Thomas_2019_ICCV,
author = {Thomas, Hugues and Qi, Charles R. and Deschaud, Jean-Emmanuel and Marcotegui, Beatriz and Goulette, Francois and Guibas, Leonidas J.},
title = {{KPConv}: Flexible and Deformable Convolution for Point Clouds},
booktitle = {Proceedings of the IEEE/CVF International Conference on Computer Vision (ICCV)},
month = {October},
year = {2019}
}

@inproceedings{NEURIPS2022_d78ece66,
 author = {Wu, Xiaoyang and Lao, Yixing and Jiang, Li and Liu, Xihui and Zhao, Hengshuang},
 booktitle = {Advances in Neural Information Processing Systems},
 editor = {S. Koyejo and S. Mohamed and A. Agarwal and D. Belgrave and K. Cho and A. Oh},
 pages = {33330--33342},
 publisher = {Curran Associates, Inc.},
 title = {{Point Transformer V2}: Grouped Vector Attention and Partition-based Pooling},
 url = {https://proceedings.neurips.cc/paper_files/paper/2022/file/d78ece6613953f46501b958b7bb4582f-Paper-Conference.pdf},
 volume = {35},
 year = {2022}
}

@inproceedings{ronneberger2015unetconvolutionalnetworksbiomedical,
  title={{U-Net}: Convolutional Networks for Biomedical Image Segmentation},
  author={Olaf Ronneberger and Philipp Fischer and Thomas Brox},
  booktitle={Medical Image Computing and Computer-Assisted Intervention -- MICCAI 2015},
  series={Lecture Notes in Computer Science},
  volume={9351},
  pages={234--241},
  publisher={Springer},
  year={2015},
  doi={10.1007/978-3-319-24574-4_28}
}

@inproceedings{pointnet,
  title     = {{PointNet}: Deep Learning on Point Sets for {3D} Classification and Segmentation},
  author    = {Charles R. Qi and Hao Su and Kaichun Mo and Leonidas J. Guibas},
  booktitle = {Proceedings of the IEEE Conference on Computer Vision and Pattern Recognition (CVPR)},
  year      = {2017},
  doi       = {10.1109/CVPR.2017.16},
  url       = {https://arxiv.org/abs/1612.00593}
}

@inproceedings{2drope,
  title={Rotary Position Embedding for Vision Transformer},
  author={Byeongho Heo and Song Park and Dongyoon Han and Sangdoo Yun},
  booktitle={European Conference on Computer Vision (ECCV)},
  year={2024},
  pages={289--305},
  doi={10.1007/978-3-031-72684-2_17},
  url={https://arxiv.org/abs/2403.13298}
}

@misc{pi05,
      title={$\pi_{0.5}$: a Vision-Language-Action Model with Open-World Generalization}, 
      author={{Physical Intelligence} and Kevin Black and Noah Brown and James Darpinian and Karan Dhabalia and Danny Driess and Adnan Esmail and Michael Equi and Chelsea Finn and Niccolo Fusai and Manuel Y. Galliker and Dibya Ghosh and Lachy Groom and Karol Hausman and Brian Ichter and Szymon Jakubczak and Tim Jones and Liyiming Ke and Devin LeBlanc and Sergey Levine and Adrian Li-Bell and Mohith Mothukuri and Suraj Nair and Karl Pertsch and Allen Z. Ren and Lucy Xiaoyang Shi and Laura Smith and Jost Tobias Springenberg and Kyle Stachowicz and James Tanner and Quan Vuong and Homer Walke and Anna Walling and Haohuan Wang and Lili Yu and Ury Zhilinsky},
      year={2025},
      eprint={2504.16054},
      archivePrefix={arXiv},
      primaryClass={cs.LG},
      url={https://arxiv.org/abs/2504.16054}, 
}

@inproceedings{3dda,
  title={{3D Diffuser Actor}: Policy Diffusion with {3D} Scene Representations},
  author={Tsung-Wei Ke and Nikolaos Gkanatsios and Katerina Fragkiadaki},
  booktitle={Conference on Robot Learning (CoRL)},
  year={2024},
  pages={1949--1974},
  url={https://proceedings.mlr.press/v270/ke25a.html}
}

@article{dinov2,
  title   = {{DINOv2}: Learning Robust Visual Features without Supervision},
  author  = {Maxime Oquab and Timoth{\'e}e Darcet and Th{\'e}o Moutakanni and Huy V. Vo and Marc Szafraniec and Vasil Khalidov and Pierre Fernandez and Daniel Haziza and Francisco Massa and Alaaeldin El-Nouby and Mahmoud Assran and Nicolas Ballas and Wojciech Galuba and Russell Howes and Po-Yao Huang and Shang-Wen Li and Ishan Misra and Michael Rabbat and Vasu Sharma and Gabriel Synnaeve and Hu Xu and Herv{\'e} Jegou and Julien Mairal and Patrick Labatut and Armand Joulin and Piotr Bojanowski},
  journal = {Transactions on Machine Learning Research},
  issn    = {2835-8856},
  year    = {2024},
  url     = {https://openreview.net/forum?id=a68SUt6zFt}
}

@article{deng2025stereovla,
 title={StereoVLA: Enhancing Vision-Language-Action Models with Stereo Vision},
 author={Deng, Shengliang and Yan, Mi and Zheng, Yixin and Su, Jiayi and Zhang, Wenhao and Zeng, Yitao and Zhao, Xiaoguang and Cui, Heming and Zhang, Zhizheng and Wang, He},
 journal={arXiv preprint arXiv:2512.21970}, year={2025}, url={https://arxiv.org/abs/2512.21970}
}

@article{li2025pointvla,
 title={PointVLA: Injecting the 3D World into Vision-Language-Action Models},
 author={Li, Chengmeng and Wen, Junjie and Peng, Yan and Peng, Yaxin and Feng, Feifei and Zhu, Yichen},
 journal={arXiv preprint arXiv:2503.07511}, year={2025}, url={https://arxiv.org/abs/2503.07511}
}

@inproceedings{li2025bridgevla,
  title     = {{BridgeVLA}: Input-Output Alignment for Efficient {3D} Manipulation Learning with Vision-Language Models},
  author    = {Peiyan Li and Yixiang Chen and Hongtao Wu and Xiao Ma and Xiangnan Wu and Yan Huang and Liang Wang and Tao Kong and Tieniu Tan},
  booktitle = {Advances in Neural Information Processing Systems (NeurIPS)},
  year      = {2025},
  url       = {https://openreview.net/forum?id=ffBF6hYuQv}
}

@inproceedings{wen2026fastfoundationstereo,
  title     = {{Fast-FoundationStereo}: Real-Time Zero-Shot Stereo Matching},
  author    = {Bowen Wen and Shaurya Dewan and Stan Birchfield},
  booktitle = {Proceedings of the IEEE/CVF Conference on Computer Vision and Pattern Recognition (CVPR)},
  pages     = {7513--7524},
  year      = {2026},
  eprint    = {2512.11130},
  archivePrefix = {arXiv}
}

@inproceedings{qu2025spatialvla,
  title     = {{SpatialVLA}: Exploring Spatial Representations for Visual-Language-Action Model},
  author    = {Delin Qu and Haoming Song and Qizhi Chen and Yuanqi Yao and Xinyi Ye and Yan Ding and Zhigang Wang and JiaYuan Gu and Bin Zhao and Dong Wang and Xuelong Li},
  booktitle = {Robotics: Science and Systems (RSS)},
  year      = {2025},
  eprint    = {2501.15830},
  archivePrefix = {arXiv}
}

@misc{yang2025fp3,
  title     = {{FP3}: A {3D} Foundation Policy for Robotic Manipulation},
  author    = {Rujia Yang and Geng Chen and Chuan Wen and Yang Gao},
  year      = {2025},
  eprint    = {2503.08950},
  archivePrefix = {arXiv},
  primaryClass  = {cs.RO},
  url       = {https://arxiv.org/abs/2503.08950}
}

@inproceedings{gervet2023act3d,
  title     = {{Act3D}: {3D} Feature Field Transformers for Multi-Task Robotic Manipulation},
  author    = {Theophile Gervet and Zhou Xian and Nikolaos Gkanatsios and Katerina Fragkiadaki},
  booktitle = {Proceedings of The 7th Conference on Robot Learning (CoRL)},
  series    = {Proceedings of Machine Learning Research},
  volume    = {229},
  pages     = {3949--3965},
  year      = {2023}
}

@inproceedings{goyal2024rvt2,
  title     = {{RVT-2}: Learning Precise Manipulation from Few Demonstrations},
  author    = {Ankit Goyal and Valts Blukis and Jie Xu and Yijie Guo and Yu-Wei Chao and Dieter Fox},
  booktitle = {Robotics: Science and Systems (RSS)},
  year      = {2024},
  eprint    = {2406.08545},
  archivePrefix = {arXiv}
}

@article{li2026mvvdp,
  title   = {Multi-View Video Diffusion Policy: A {3D} Spatio-Temporal-Aware Video Action Model},
  author  = {Peiyan Li and Yixiang Chen and Yuan Xu and Jiabing Yang and Xiangnan Wu and Jun Guo and Nan Sun and Long Qian and Xinghang Li and Xin Xiao and Jing Liu and Nianfeng Liu and Tao Kong and Yan Huang and Liang Wang and Tieniu Tan},
  journal = {arXiv preprint arXiv:2604.03181},
  year    = {2026},
  url     = {https://arxiv.org/abs/2604.03181}
}

@inproceedings{wang2025gaudp,
  title     = {{GauDP}: Reinventing Multi-Agent Collaboration through Gaussian-Image Synergy in Diffusion Policies},
  author    = {Ziye Wang and Li Kang and Yiran Qin and Jiahua Ma and Zhanglin Peng and Lei Bai and Ruimao Zhang},
  booktitle = {Advances in Neural Information Processing Systems (NeurIPS)},
  year      = {2025},
  eprint    = {2511.00998},
  archivePrefix = {arXiv}
}
